\pdfoutput=1
\documentclass{article}

\def\includehome{./include}

\PassOptionsToPackage{numbers, compress}{natbib}
\usepackage[preprint]{neurips_2020}

\usepackage[utf8]{inputenc} 
\usepackage[T1]{fontenc}    
\usepackage{hyperref}       
\usepackage{url}            
\usepackage{booktabs}       
\usepackage{amsfonts}       
\usepackage{nicefrac}       
\usepackage{microtype}      
\usepackage{mathtools}

\usepackage{xcolor}
\usepackage{dsfont}
\usepackage{graphicx}
\usepackage{grffile}
\usepackage{multirow}
\usepackage{subcaption}
\usepackage{tikz}
\usepackage{enumitem}
\usepackage{bm}
\usepackage{amsmath}
\usepackage{amssymb}
\usepackage{amsthm}
\usepackage{pifont}
\usepackage{animate}
\usepackage{footnote}
\usepackage{tablefootnote}
\usepackage{makecell}
\usepackage{floatrow}
\newfloatcommand{capbtabbox}{table}[][\FBwidth]

\makesavenoteenv{tabular}

\usepackage[ruled,lined,longend]{algorithm2e}
\definecolor{darkblue}{rgb}{0.0,0.0,0.5}

\SetCommentSty{mycommfont}
\usepackage{wrapfig}
\usepackage{tabularx}

\newlength\mylen
\newcommand\myinput[1]{%
  \settowidth\mylen{\KwIn{}}%
  \setlength\hangindent{\mylen}%
  \hspace*{\mylen}#1\\}

\newcommand{\scalarSup}[2]{{#1}(#2)}

\newcommand{\vectorSup}[2]{\bm{#1}{(#2)}}

\newcommand{\mymatrix}[1]{\bm{#1}}
	
\newcommand{\matrixSup}[2]{\bm{#1}(#2)}

\newcommand{\tensor}[1]{\mathcal{#1}}
\newcommand{\mytensor}[1]{\mathcal{#1}}

\newcommand{\tensorSup}[2]{\mathcal{#1}(#2)}
\newcommand{\tensorInd}[3]{\mathcal{#1}(#3)_{#2}}

\newcommand{\RNN}{RNN\xspace}
\newcommand{\RNNlong}{recurrent neural network\xspace}

\newcommand{\LSTM}{LSTM}

\newcommand{\LSTMLONG}{Long Short-Term Memory\xspace}

\newcommand{\ConvLSTM}{ConvLSTM\xspace}
\newcommand{\ConvLSTMlong}{convolutional LSTM\xspace}

\newcommand{\ConvTTLSTM}{Conv-TT-LSTM\xspace}

\newcommand{\ConvTTLSTMLONG}{Convolutional Tensor-Train LSTM\xspace}

\newcommand{\TTD}{TTD\xspace}
\newcommand{\TTDlong}{tensor-train decomposition\xspace}

\newcommand{\CTTD}{CTTD\xspace}
\newcommand{\CTTDlong}{convolutional tensor-train decomposition\xspace}

\title{Convolutional Tensor-Train LSTM for Spatio-Temporal Learning}

\makeatletter
\newcommand{\printfnsymbol}[1]{%
  \textsuperscript{\@fnsymbol{#1}}%
}
\makeatother
\author{
Jiahao Su\thanks{Equal contribution}$^{\;\;12}$\thanks{This work was done while the author was an intern at NVIDIA.} \hskip2em Wonmin Byeon\printfnsymbol{1}$^{2}$ \hskip2em Jean Kossaifi$^2$ \\
\texttt{\small  jiahaosu@umd.edu, \{wbyeon,jkossaifi\}@nvidia.com} \\ 
\And
Furong Huang$^1$ \hskip2em Jan Kautz$^2$ \hskip2em Anima Anandkumar$^2$ \\
\texttt{\small furongh@cs.umd.edu, \{jkautz,aanandkumar\}@nvidia.com}\\ \\
$^1$University of Maryland, College Park, MD \hskip1em $^2$NVIDIA Research, Santa Clara, CA \\
}

\newcommand\blfootnote[1]{%
  \begingroup
  \renewcommand\thefootnote{}\footnote{#1}%
  \addtocounter{footnote}{-1}%
  \endgroup
}

\begin{document}
\maketitle

\begin{abstract}
Learning from spatio-temporal data has numerous applications such as human-behavior analysis, object tracking, video compression, and physics simulation.
However, existing methods still perform poorly on challenging video tasks such as long-term forecasting.
This is because these kinds of challenging tasks require learning long-term spatio-temporal correlations in the video sequence.
In this paper, we propose a higher-order convolutional LSTM model that can efficiently learn these correlations, along with a succinct representations of the history. 
This is accomplished through a novel tensor-train module that performs prediction by combining convolutional features across time. 
To make this feasible in terms of computation and memory requirements, we propose a novel convolutional tensor-train decomposition of the higher-order model. 
This decomposition reduces the model complexity by jointly approximating a sequence of convolutional kernels as a low-rank tensor-train factorization.
As a result, our model outperforms existing approaches, but uses only a fraction of parameters, including the baseline models.
Our results achieve state-of-the-art performance in a wide range of applications and datasets, including the multi-steps video prediction on the Moving-MNIST-2 and KTH action datasets as well as early activity recognition on the Something-Something V2 dataset. \blfootnote{Project page: \url{https://sites.google.com/nvidia.com/conv-tt-lstm}}
\end{abstract}

\section{Introduction}
\label{sec:introduction}

While computer vision has achieved remarkable successes, e.g., on image classification, many real-life tasks remain out-of-reach for current deep learning systems, such as prediction from complex spatio-temporal data.
This naturally arises in a wide range of applications such as autonomous driving, robot control~\citep{finn2017deep}, visual perception tasks such as action recognition~\citep{srivastava2015unsupervised} or object tracking~\citep{alahi2016social}, and even weather prediction~\citep{xingjian2015convolutional}.
This kind of video understanding problems is challenging, 
since they require learning spatial-temporal representations
that capture both content and dynamics simultaneously.

\textbf{Learning from (video) sequences.} 
Most state-of-the-art video models are based on {\RNNlong}s~({\RNN}s), typically some variations of {\em Convolutional LSTM} {(\ConvLSTM)} where spatio-temporal information is encoded explicitly in each cell~\citep{xingjian2015convolutional,wang2017predrnn, wang2018predrnnpp,wang2018eidetic}.
These {\RNN}s are first-order Markovian models in nature, meaning that the hidden states are updated using information from the previous time step only, resulting in an intrinsic difficulty in capturing long-range temporal correlations.

\textbf{Incorporating higher-order correlations.}
For one-dimensional sequence modeling, higher-order generalizations of {\RNN}s have previously been proposed for long-term forecasting problems~\citep{soltani2016higher,yu2017long}.
Higher-order {\RNN}s explicitly incorporate a longer history of previous states in each update. This requires higher-order tensors to characterize the transition function (instead of a transition matrix as in first-order {\RNN}s).
However, this typically leads to an exponential blow-up in the complexity of the transition function.
This problem is further compounded when trying to generalize {\ConvLSTM} to higher-orders and these generalizations have not been explored.

\textbf{Scaling up with tensor methods.}
To avoid the exponential blow-up in the complexity of transition function,
tensor decompositions~\citep{anandkumar2014tensor}
have been investigated within higher-order RNNs~\citep{yu2017long}.
Tensor decomposition not only avoids the exponential growth of model complexity, but also introduces an information bottleneck that facilitates effective representation learning.
This restricts how much information can be passed on from one sub-system to another in a learning system~\cite{tishby2015deep,achille2018emergence}.
Previously, low-rank tensor factorization has been used
to improve a variety of deep network architectures~\cite{lebedev2014speeding,kim2015compression,kossaifi2019t,kossaifi2019efficient}.
However, it has not been analyzed in the context of spatio-temporal {\LSTM}s.
The only approach that leveraged tensor factorization for compact higher-order {\LSTM}s~\citep{yu2017long} considers exclusively sequence forecasting
and cannot be directly extended to general spatio-temporal data.

\textbf{Generalizing ConvLSTM to higher-orders.} When extending to higher-orders, we aim to design a transition function that is able to leverage all previous hidden states and satisfies three properties: \textbf{(i)} the spatial structure in the hidden states is preserved; \textbf{(ii)} the receptive field increases with time. In other words, the longer the temporal correlation captured, the larger the spatial context should be. \textbf{(iii)} Finally, space and time complexities grow at most linearly with the number of times steps.
Because previous transition functions in higher-order RNNs were designed specifically for one-dimensional sequence, when directly extended to spatio-temporal data they do not satisfy all three properties. 
A direct extension fails to preserves the spatial stricture or increases the complexity exponentially. 


\textbf{Contributions.}
In this paper, we propose a higher-order Convolutional LSTM model for complex spatio-temporal data satisfying all three properties.
Our model incorporates a long history of states in each update, while preserving their spatial structure using convolutional operations. 
Directly constructing such a model leads to an exponential growth of parameters in both spatial and temporal dimensions.
Instead, our model is made computationally tractable via a novel convolutional tensor-train decomposition, which recursively performs a convolutional factorization of the kernels across time.
In addition to the parameter reduction, this low-rank factorization introduces an information bottleneck that helps to learn better representations. 
As a result, it achieves better results than previous works with only a fraction of parameters.

We empirically demonstrate the performance of our model on several challenging tasks, including early activity recognition and video prediction.
We report an absolute increase of 8\% in accuracy over the state-of-the-art~\citep{wang2018eidetic} for early activity recognition on the Something-Something v2 dataset. 
Our model outperforms both 3D-CNN and ConvLSTM by a large margin. We also report a new state-of-the-art for multi-step video prediction on both Moving-MNIST-2 and KTH datasets.

Finally, we propose a principled procedure to train higher-order models: we design a preprocessing module to incorporate longer temporal context and highlight the importance of appropriate gradient clipping and learning scheduling to improve training of higher-order models. We train all models with these strategies and report consistent improvements in performance.
\section{Background: Convolutional LSTM and Higher-order LSTM}
\label{sec:background}
In this section, we briefly review {\em {\LSTMLONG}} (LSTM), and its generalizations
{\em Convolutional LSTM} for sptio-temporal modeling, and {\em higher-order LSTM} for  learning long-term dynamics.

{\bf {\LSTMLONG} (\LSTM)~\citep{hochreiter1997long}} is a first-order Markovian model widely used in 1D sequence learning. At each time step, an {\LSTM} cell updates its states \(\{\vectorSup{h}{t}, \vectorSup{c}{t}\}\) using the immediate previous states \(\{\vectorSup{h}{t-1}, \vectorSup{c}{t-1}\}\) and the current input \(\vectorSup{x}{t}\) as
\begin{subequations}
\begin{gather}
[\vectorSup{i}{t}; \vectorSup{f}{t};
\vectorSup{\tilde{c}}{t}; \vectorSup{o}{t}]
= \sigma(\mymatrix{W} \vectorSup{x}{t} + \mymatrix{K} \vectorSup{h}{t-1}) 
\label{eq:lstm-step-1}; \\
\vectorSup{c}{t} = \vectorSup{c}{t-1} \circ \vectorSup{f}{t} + \vectorSup{\tilde{c}}{t} \circ \vectorSup{i}{t}; ~
\vectorSup{h}{t} = \vectorSup{o}{t} \circ \sigma(\vectorSup{c}{t}),
\label{eq:lstm-step-2}
\end{gather}
\end{subequations}
where \(\sigma(\cdot)\) denotes a \(\mathsf{sigmoid}(\cdot)\) applied to the {\em input gate} \(\vectorSup{i}{t}\), {\em forget gate} \(\vectorSup{f}{t}\) and {\em output gate} \(\vectorSup{o}{t}\),  and a \(\tanh(\cdot)\) applied to the {\em memory cell} \(\vectorSup{\tilde{c}}{t}\) and {\em cell state} \(\vectorSup{{c}}{t}\). \(\circ\) denotes element-wise product. 
{\LSTM}s have two major restrictions: {\bf(a)} only 1D-sequences can be modeled, not spatio-temporal data such as videos; {\bf(b)} they are difficult to capture long-term dynamics as first-order models.

{\bf Convolutional LSTM ({\ConvLSTM})~\citep{xingjian2015convolutional}} addresses the limitation {\bf(a)} by extending {\LSTM} to model spatio-temporal structures within each cell, i.e.\ the states, cell memory, gates and parameters are all encoded as high-dimensional tensors. Furthermore, Eq.~\eqref{eq:lstm-step-1} is replaced by
\begin{equation}
[\tensorSup{I}{t}; \tensorSup{F}{t}; 
\tensorSup{\Tilde{C}}{t}; \tensorSup{O}{t} ]
= \sigma(\tensor{W} \ast \tensorSup{X}{t}
+ \tensor{K} \ast \tensorSup{H}{t-1}),
\label{eq:convlstm}
\end{equation}
where \(\ast\) defines convolution between states and parameters as in convolutional neural networks.

\textbf{Higher-order LSTM (HO-LSTM)} is a higher-order Markovian generalization of the basic {\LSTM}, which partially addresses the limitation {\bf(b)} in modeling long-term dynamics. Specifically, HO-LSTM explicitly incorporates more previous states in each update, replacing the first step in {\LSTM } by
\begin{equation}
\left[ \vectorSup{i}{t}; \vectorSup{f}{t};
\vectorSup{\tilde{c}}{t}; \vectorSup{o}{t} \right]
= \sigma\left(\mymatrix{W} \vectorSup{x}{t} 
+ \Phi \left( \vectorSup{h}{t-1}, \cdots, \vectorSup{h}{t-N}\right) \right),
\label{eq:higher-order-lstm}
\end{equation}
where \(\Phi\) combines \(N\) previous states \(\{\vectorSup{h}{t-1}, \cdots, \vectorSup{h}{t-N}\}\) and \(N\) is the {\em order} of the {HO-LSTM}.
Two realizations of \(\Phi\) have been proposed: a linear function~\citep{soltani2016higher} and a polynomial one~\citep{yu2017long}:
\begin{align}
\text{Linear: } & \quad
\Phi \left( \vectorSup{h}{t-1}, \cdots, \vectorSup{h}{t-M}; ~ \matrixSup{T}{1}, \cdots, \matrixSup{T}{N} \right) = \sum\nolimits_{i = 1}^{N} \matrixSup{T}{i} \vectorSup{h}{t-i}.
\label{eq:higher-order-linear} \\
\text{Polynomial: } & \quad
\Phi \left( \vectorSup{h}{t-1}, \cdots, \vectorSup{h}{t-N}; ~\tensor{T} \right) = \left< \tensor{T}, ~ \vectorSup{h}{t-1} \otimes \cdots \otimes \vectorSup{h}{t-N} \right>.
\label{eq:higher-order-polynomial}
\end{align}
While a linear function requires the numbers of parameters and operations growing linearly in \(N\), a polynomial function has space/computational complexity exponential in \(N\) if implemented naively.
\section{Methodology: Convolutional Tensor-Train LSTM}
\label{sec:conv-tt-lstm}

Here, we detail the challenges and requirements for designing a higher-order {\ConvLSTM}. We then introduce our model, and motivate the design of each module by these requirements.

\subsection{Extending  {\ConvLSTM} to Higher-orders}
\label{sub:higher-order-convlstm}

We can express a general higher-order {\ConvLSTM} by combining several previous states when computing the gates for each step:
\begin{equation}
\big[ \tensorSup{I}{t}; \tensorSup{F}{t};  
\vectorSup{\tilde{C}}{t}; \tensorSup{O}{t} \big]
= \sigma\left(\mytensor{W} \ast \tensorSup{X}{t} 
+ \Phi \left( \tensorSup{H}{t-1}, \cdots, \tensorSup{H}{t-N}\right) \right).
\label{eq:higher-order-convlstm}
\end{equation}
The choice of a suitable function \(\Phi\) for a spatio-temporal learning problem, however, is difficult, as it should satisfy the following properties: 
\begin{enumerate}[label={(\arabic*)}, leftmargin=*, itemsep=0pt, topsep=0pt]
    \item The spatial structure in the hidden states \(\tensorSup{H}{t}\)'s is preserved by the operations in \(\Phi\).
    \item The size of the receptive field for \(\tensorSup{H}{t-i}\) increases with \(i\), the time gap from the current step (\(i = 1, 2, \cdots, N\)). In other words, the longer temporal correlation captured, the larger the considered spatial context should be.
    \item Both space and time complexities grow {\em at most} linearly with times steps \(N\), i.e.\ \(O(N)\).
\end{enumerate}

\textbf{Limitations of previous approaches.} While it is possible to construct a function \(\Phi\) by extending the linear function in Eq.\eqref{eq:higher-order-linear} or the polynomial function in Eq.\eqref{eq:higher-order-polynomial} to the tensor case, none of these extensions satisfy the all three properties.
While the polynomial function with {\TTDlong}~\citep{yu2017long} meets requirement {\bf (3)}, the operations do not preserve the spatial structures in the hidden states. 
On the other hand, augmenting the linear function with convolutions leads to a function:
\begin{equation}
\Phi \left( \tensorSup{H}{t-1}, \cdots, \tensorSup{H}{t-N}; ~ \tensorSup{K}{1}, \cdots, \tensorSup{K}{N} \right) = \sum\nolimits_{i = 1}^{N} \tensorSup{K}{i} \ast \tensorSup{H}{t - i} 
\label{eq:higher-order-convolutional}
\end{equation}
which does not satisfy requirement {\bf(2)} if all \(\tensorSup{K}{i}\) contain filters of the same size \(K\). An immediate remedy is to expand \(\tensorSup{K}{i}\) such that its filter size \(\scalarSup{K}{i}\) grows linearly in \(i\). However, the resulting function requires \(O(N^3)\) space/computational complexity, violating the requirement {\bf(3)}.

\begin{figure}[!htbp]
  \includegraphics[trim={0.5cm 0.5cm 0.5cm 0.5cm},clip, width= 0.90\linewidth]{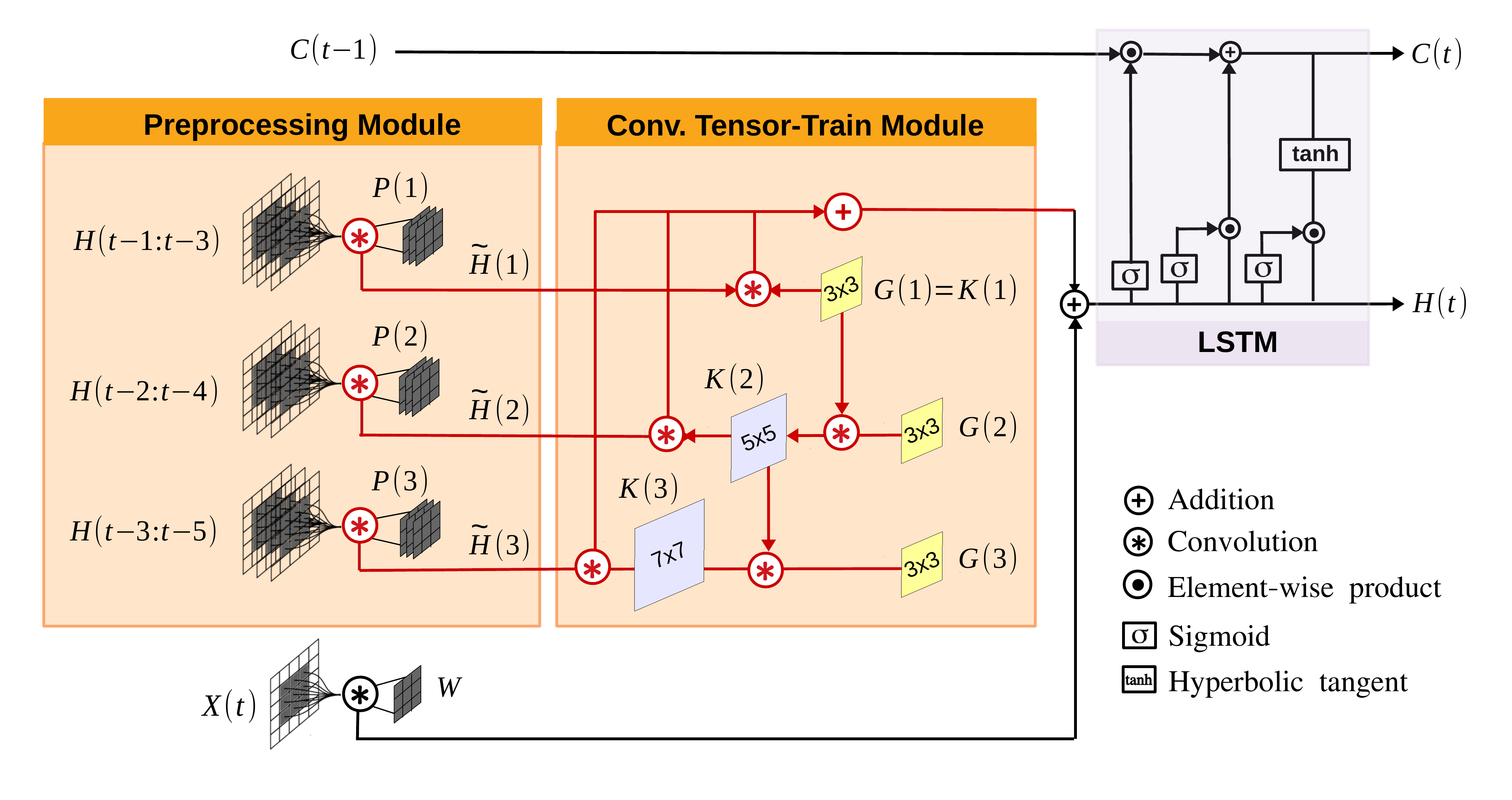}
  \caption{{\bf {\ConvTTLSTMLONG}}. The {\em preprocessing module} first groups the previous hidden states into overlapping sets with a sliding window, and reduces the number of channels in each group using a convolutional layer. The {\em convolutional tensor-train module} takes the results, aggregates their spatio-temporal information, and computes the gates for the {\LSTM} update. 
  The diagram visualizes a {\ConvTTLSTM} with one channel. When {\ConvTTLSTM} has multiple channels, the addition also accumulates the results from multiple channels.}
  \label{fig:convolutional-tensor-train}
\end{figure}

\subsection{Designing an Effective and Efficient Higher-order {\ConvLSTM}}
\label{sub:conv-tt-lstm}
In order to satisfy all three requirements {\bf (1)-(3)} introduced above, and enable efficient learning/inference, we propose a novel {\em \CTTDlong} (\CTTD) that leverages a tensor-train structure~\citep{oseledets2011tensor} to jointly express the convolutional kernels \(\{\tensorSup{K}{1}, \cdots, \tensorSup{K}{N}\}\) in Eq.\eqref{eq:higher-order-convolutional} as a series of smaller factors \(\{\tensorSup{G}{1}, \cdots, \tensorSup{G}{N}\}\) while maintaining their spatial structures.

\textbf{Convolutional Tensor-Train module.} 
Concretely, let \(\tensorSup{K}{i}\) be the \(i\)-th kernel in Eq.\eqref{eq:higher-order-convolutional}, of size \([\scalarSup{K}{i} \times \scalarSup{K}{i} \times \scalarSup{C}{i} \times \scalarSup{C}{0}]\), where \(\scalarSup{K}{i} = i [\scalarSup{K}{1} - 1] + 1\) is the filter size that increases linearly with \(i\); \(\scalarSup{K}{1}\) is the initial filter size; \(\scalarSup{C}{i}\) is the number of channels in \(\tensorSup{H}{t-i}\); and \(\scalarSup{C}{0}\) is the number of channels for the output of the function \(\Phi\) (thus \(\scalarSup{C}{0} = 4 \times C_\text{out}\), where \(C_\text{out}\) is the number of channels of the higher-order \ConvLSTM).
The \CTTD factorizes \(\tensorSup{K}{i}\) using a subset of factors \(\{\tensorSup{G}{1}, \cdots, \tensorSup{G}{i}\}\) up to index \(i\) such that
{\small
\begingroup
\begin{equation}
\tensorInd{K}{\bm{:}, \bm{:}, c_{i}, c_0}{i} \triangleq \mathsf{CTTD}\left(\{\tensorSup{G}{j}\}_{j = 1}^{i} \right) 
= \sum_{c_{i - 1} = 1}^{\scalarSup{C}{i-1}} \cdots \sum_{c_1 = 1}^{\scalarSup{C}{1}} \tensorInd{G}{\bm{:}, \bm{:}, c_{i}, c_{i-1}}{i} \ast \cdots \ast \tensorInd{G}{\bm{:}, \bm{:}, c_2, c_1}{2} \ast \tensorInd{G}{\bm{:}, \bm{:}, c_1, c_0}{1},
\label{eq:CTTD}
\end{equation}
\endgroup}%
where \(\tensorSup{G}{i}\) has size \([\scalarSup{K}{1} \times \scalarSup{K}{1} \times \scalarSup{C}{i} \times \scalarSup{C}{i-1}]\). 
The number of factors \(N\) is known as the {\em order of the decomposition}, and the {\em ranks of the decomposition} \(\{\scalarSup{C}{1}, \cdots, \scalarSup{C}{N-1}\}\) are the channels of the convolutional kernels.

Notice that the same set of factors \(\{\tensorSup{G}{1}, \cdots, \tensorSup{G}{N}\}\) is reused to construct all convolutional kernels \(\{\tensorSup{K}{1}, \cdots, \tensorSup{K}{N}\}\), such that the number of total parameters grows linearly in \(N\). 
In fact, the convolutional kernel \(\tensorSup{K}{i+1}\) can be recursively constructed as \(\tensorSup{K}{i} = \tensorSup{G}{i} \ast \tensorSup{K}{i-1} \) with \(\tensorSup{K}{1} = \tensorSup{G}{1}\) and \(\tensorInd{K}{\bm{:}, \bm{:}, c_i, c_0}{i} = \sum_{c_{i-1}} \tensorInd{G}{\bm{:}, \bm{:}, c_{i}, c_{i-1}}{i} \ast \tensorInd{K}{\bm{:}, \bm{:}, c_{i-1}, c_0}{i-1}\) for \(i \geq 2\).

This results into in a \textit{convolutional tensor-train module} that we use for function $\Phi$ in Eq.\eqref{eq:higher-order-convolutional}:
{\small
\begingroup
\begin{equation}
\Phi = \mathsf{CTT} ( \tensorSup{H}{t-1}, \cdots, \tensorSup{H}{t-N}; ~\tensorSup{G}{1}, \cdots, \tensorSup{G}{N} ) = \sum\nolimits_{i = 1}^{N} \mathsf{CTTD} \big( \{\tensorSup{G}{j}\}_{j = 1}^{i} \big) \ast \tensorSup{H}{t-i}
\label{eq:conv-tt-lstm-v1}
\end{equation}
\endgroup}%
In \autoref{app:tensor-trains}, we show that the computation of Eq.\eqref{eq:conv-tt-lstm-v1} can be done in linear time \(O(N)\), thus the construction of \(\mathsf{CTT}\) satisfies all requirements \textbf{(1)-(3)}.

\textbf{Preprocessing module.}
In Eq.\eqref{eq:conv-tt-lstm-v1}, we use the raw hidden states \(\tensorSup{H}{t}\) as inputs to \(\mathsf{CTT}\). This design has two limitations: {\bf(a)} The number of past steps in \(\mathsf{CTT}\) (i.e. the order of the higher-order {\ConvLSTM}) is equal to the number of factors in {\CTTD} (i.e. the order of the tensor decomposition), which both equal to \(N\). It is prohibitive to use a long history, as a large tensor order leads to gradient vanishing/exploding problem in computing Eq.\eqref{eq:conv-tt-lstm-v1}; {\bf(b)} All the ranks \(\scalarSup{C}{i}\) are equal to the number of channels in \(\tensorSup{H}{t}\), which prevents the use of lower-ranks to further reduce the model complexity.

To address both issues, we develop a preprocessing module to reduce both the number of steps and channels in previous hidden states before they are fed into \(\mathsf{CTT}\).
Suppose the number of steps \(M\) is no less than the tensor order \(N\) (i.e.\ \(M \geq N \)),
the preprocessing collects the neighboring steps with a sliding window and reduce it into an intermediate result with \(\scalarSup{C}{i}\) channels:
\begin{equation}
\tensorSup{\Tilde{H}}{i} = \tensorSup{P}{i} \ast \left[ \tensorSup{H}{t-i}; \cdots ; \tensorSup{H}{t - i + N - M} \right]
\label{eq:sliding-window}
\end{equation}
where \(\tensorSup{P}{i}\) represents a convolutional layer that maps the concatenation \([\cdot]\) into \(\tensorSup{\Tilde{H}}{i}\).

\textbf{Convolutional Tensor-Train LSTM. } 
By combining all the above modules, we obtain our proposed {\ConvTTLSTM}, illustrated in \autoref{fig:convolutional-tensor-train} and expressed as:
\begin{equation}
\big[ \tensorSup{I}{t}; \tensorSup{F}{t};  
\vectorSup{\tilde{C}}{t}; \tensorSup{O}{t} \big]
= \sigma\big(\mytensor{W} \ast \tensorSup{X}{t} 
+ \mathsf{CTT} \big( \tensorSup{\Tilde{H}}{1}, \cdots, \tensorSup{\Tilde{H}}{N}; ~ \tensorSup{G}{1}, \cdots, \tensorSup{G}{N} \big) \big) 
\label{eq:conv-tt-lstm}
\end{equation}
This final implementation has several advantages: it drastically reduces the number of parameters and makes the higher-order {\ConvLSTM} even more compact than first-order {\ConvLSTM}. 
The low-rank constraint acts as an implicit regularizer,
leading to more generalizabled models. 
Finally, the tensor-train structure inherently encodes the correlations resulting from the natural flow of time~\cite{yu2017long}.
The full procedure can be found in \autoref{app:tensor-trains} (\autoref{alg:conv-tt-lstm-fast}).
\section{Experiments}
\label{sec:experiments}

Here, we quantitatively and empirically evaluate our approach on several datasets, for two different tasks, video prediction and early activity recognition and find that it outperforms existing approaches.

\subsection{Implementation Details}
\label{sub:implementation}

\textbf{Effective training strategy of higher-order prediction models.} 
To facilitate training, we argue for a careful choice of the learning scheduling and gradient clipping.
Specifically, various \emph{learning scheduling techniques} including learning rate decay, scheduled sampling~\citep{bengio2015scheduled} and curriculum learning with varying weighting factor are added during training. 
For video prediction, learning rate decay is used along with scheduled sampling, where scheduled sampling starts if the model does not improve for a few epochs in terms of validation loss;
For early activity recognition, learning rate decay is combined with weighting factor decay, where the weighting factor is decreased linearly \(
\lambda := \max(\lambda - \epsilon, 0)\) on plateau.
We also found \emph{gradient clipping} essential for higher-order models. 
All models are trained with ADAM optimizer~\citep{kingma2014adam}.
In the initial experiments, we found that our models are unstable at a high learning rate $1e^{-3}$, but learn poorly at a low learning rate $1e^{-4}$. Consequently, we use gradient clipping with learning rate \(1e^{-3}\),
with clipping value $1$ for all experiments.

\textbf{Evaluation.}
For \emph{video prediction}, the model predicts every pixel in the frame. 
We test our proposed models on
the KTH human action dataset~\citep{laptev2004recognizing} with resolution \(128 \times 128\) 
and on the Moving-MNIST-2 dataset~\citep{srivastava2015unsupervised} with resolution \(64 \times 64\).
All models are trained to predict \(10\) future frames given \(10\) input frames, and tested to predict \(10-40\) frames recursively. 
For \emph{early activity recognition}, we evaluate our approach on the Something-Something V2 dataset. 
Following~\cite{wang2018eidetic}, we used the subset of 41 categories defined by~\citet{goyal2017something} (Table 7). 
The prediction model is trained to predict the next \(10\) frames given \(25\% - 50\%\) of frames, and jointly classify the activity using the learned representations of the prediction model. 

\textbf{Model architecture.}
In all video prediction experiments, we use 12 RNN layers. 
For early activity recognition, we follow the base framework of \cite{wang2018eidetic}.
The prediction model consists of two layers of 2D-CNN encoder and decoder with eight RNN layers in between.  
The output of the RNN layer is fed to the classifier that contains two 2D convolutional layers and one fully-connected layer. 
We explain the detailed architecture in \autoref{app:exp}.

\textbf{Loss function.}
For video prediction, we optimize an \(\ell_1 + \ell_2\) loss \(\mathcal{L}_{\text{prediction}} = \|\mytensor{X} - \mytensor{\hat{X}}\|_F^2 + \|\mytensor{X} - \mytensor{\hat{X}}\|_1\), where \(\mytensor{X}\) and \(\mytensor{\hat{X}}\) are the ground-truth and predicted frames. For early activity recognition, we combine the prediction loss above with an additional cross entropy for classification \(\mathcal{L}_{\text{recognition}} = \lambda \cdot \mathcal{L}_{\text{prediction}} + \mathcal{L}_{\text{ce}}(y, \hat{y})\), where \(y\) and \(\hat{y}\) are the ground-truth and predicted labels. The  {\em weighting factor} \(\lambda\) balances the learning representation and exploiting the representation for activity recognition.

\textbf{Hyper-parameter selection.} 
We validate the hyper-parameters of our {\ConvTTLSTM} on though a wide grid search on the validation set. Specifically, we consider a base filter size \(S = 3, 5\), order of the decomposition \(N = 1, 2, 3, 5\), tensor ranks \(\scalarSup{C}{i} = 4, 8, 16\), and number of hidden states \(M = 1, 3, 5\). \autoref{app:exp} contains the details of our hyper-parameter search.

\begin{figure*}[b]
  \begin{subfigure}{0.47\textwidth}
  \centering
      \includegraphics[trim={0 0 0cm 0},clip,width=\textwidth]{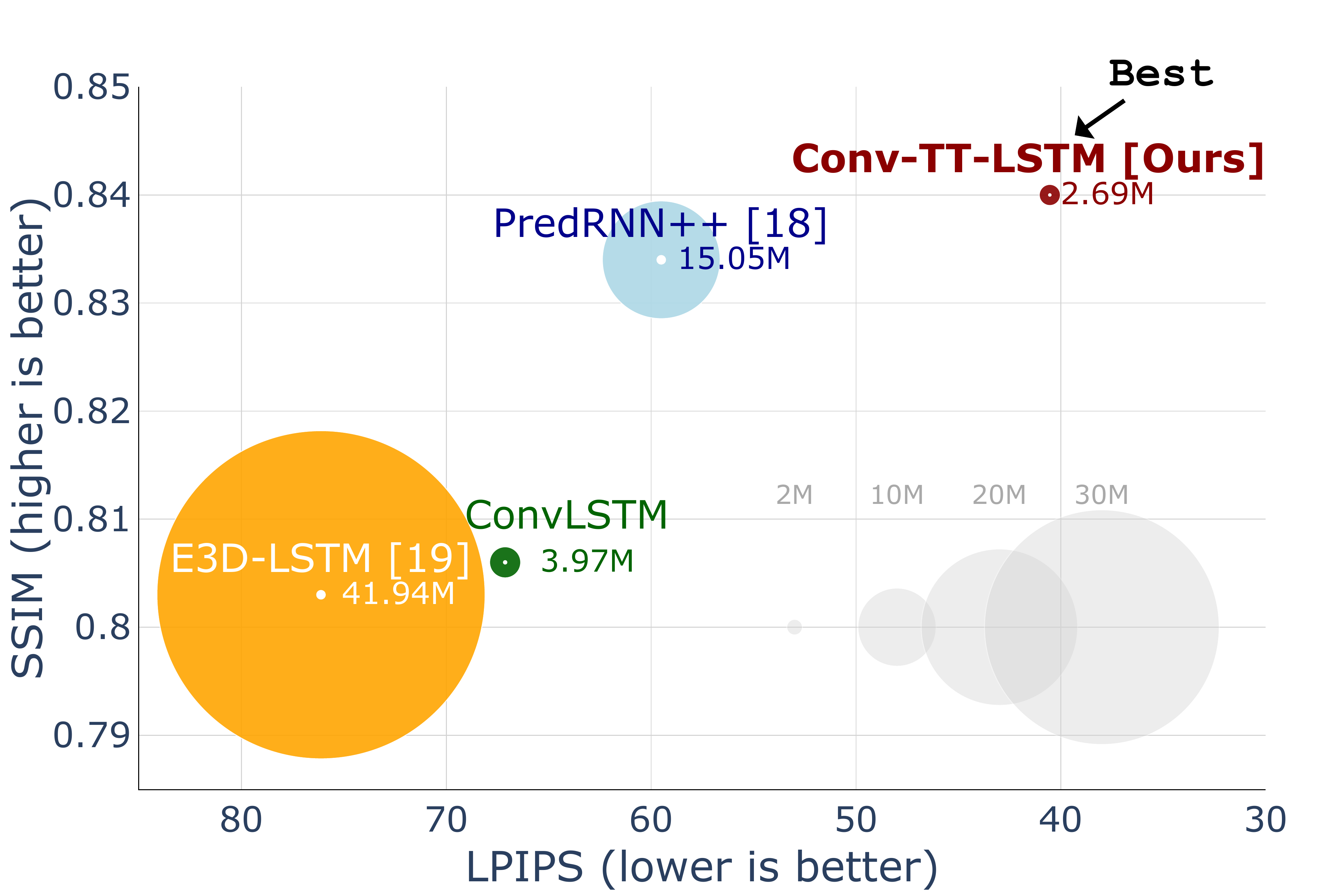}
  \end{subfigure}
  \qquad
  \begin{subfigure}{0.47\textwidth}
      \includegraphics[trim={0 0 0cm 0},clip,width=\textwidth]{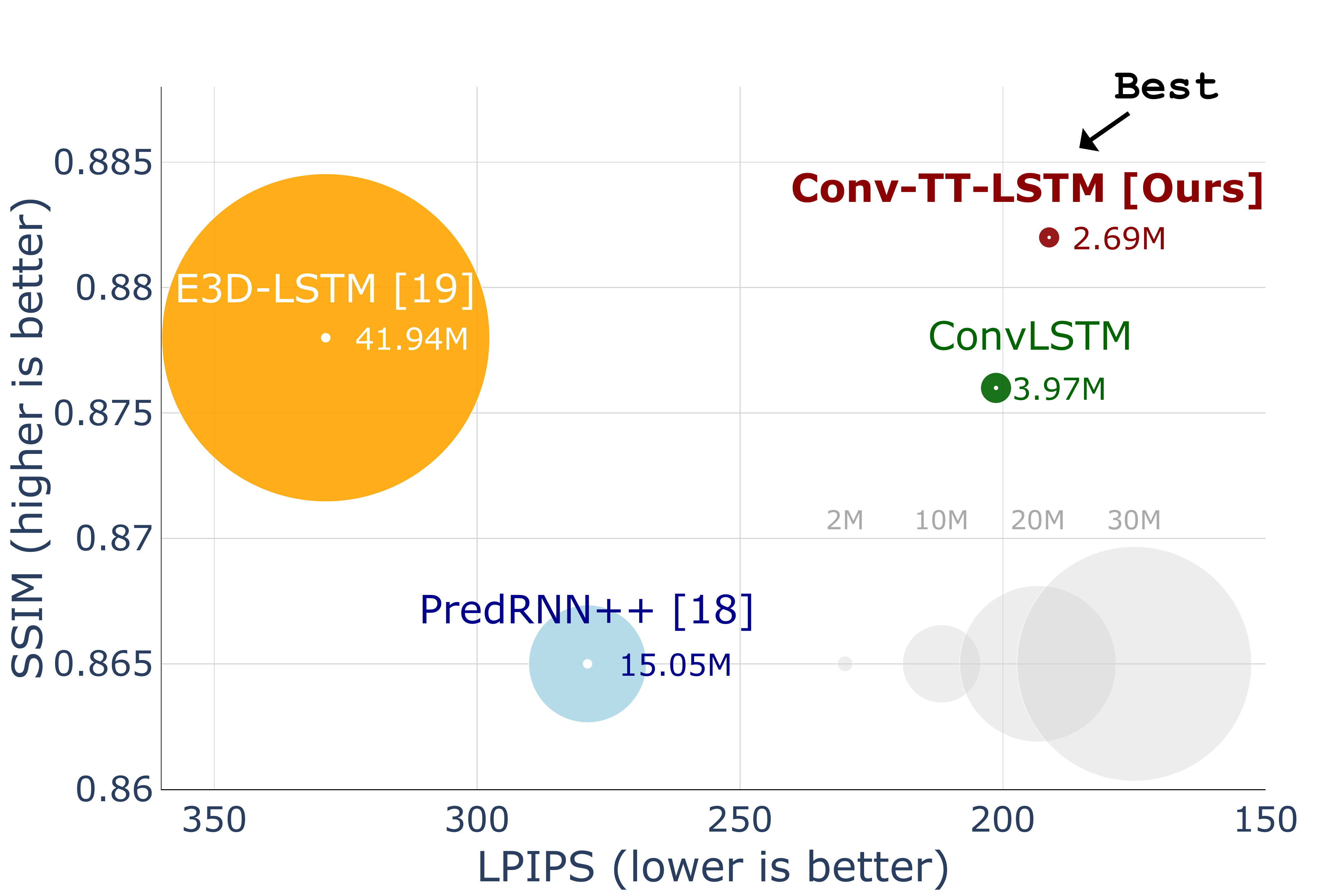}
  \end{subfigure}
  \caption{{\bf SSIM vs LPIPS scores on Moving MNIST-s2 (left) and KTH action datasets (right). The bubble size is the model size.} Higher SSIM score and lower LPIPS score are better. On both datasets and for both metrics, our approach reaches a significantly better performance than other methods while having only a fraction of the parameters.}
  \vspace{-0.3cm}
  \label{fig:ssimvslpips}
\end{figure*}

\subsection{Experimental Results}
\label{subsec:kth}

\begin{figure*}
\vspace{-0.3cm}
  \centering
  \footnotesize
  \addtolength{\tabcolsep}{-5pt}
  \begin{tabular}{c@{\hskip -0.02cm}c@{\hskip 0.07cm}c@{\hskip -0.02cm}c@{\hskip -0.02cm}c@{\hskip -0.01cm}c@{\hskip -0.01cm} c
  @{\hskip 0.4cm}c@{\hskip -0.02cm}c@{\hskip 0.07cm}c@{\hskip -0.02cm}c@{\hskip -0.02cm}c@{\hskip -0.02cm}c@{\hskip -0.02cm}c}
      \multicolumn{2}{c}{input} &
      \multicolumn{5}{c}{ground truth (top) / predictions}  & 
      \multicolumn{2}{c}{input} &
      \multicolumn{5}{c}{ground truth (top) / predictions} \\ 
      \scriptsize $t = 1$ 
      & \scriptsize $ 6$ 
      & \scriptsize $11$ 
      & \scriptsize $17$
      & \scriptsize $23$
      & \scriptsize $29$ 
      & \scriptsize $35$
      & \scriptsize $t = 1$ 
      & \scriptsize $ 6$ 
      & \scriptsize $11$ 
      & \scriptsize $15$
      & \scriptsize $19$
      & \scriptsize $23$ 
      & \scriptsize $27$ \\
      \animategraphics[height=0.932cm,loop,autoplay]{7}{figs/result_mnist_gif/ttv4_o3t3r8/inp_297_1440_}{0}{9} & 
       \includegraphics[height=0.93cm]{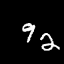} &
      \animategraphics[height=0.932cm,loop,autoplay]{7}{figs/result_mnist_gif/ttv4_o3t3r8/gt_297_1440_}{0}{29} & 
      \includegraphics[height=0.93cm]{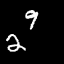} &
      \includegraphics[height=0.93cm]{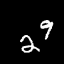} &
      \includegraphics[height=0.93cm]{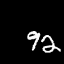} &
      \includegraphics[height=0.93cm]{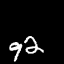} &
      \animategraphics[height=0.93cm,loop,autoplay]{8}{figs/result_kth_gif/ttv4_o3t3r8/gt_190_3296_}{0}{19} & 
      \includegraphics[height=0.93cm]{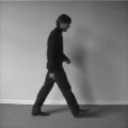} &
      \animategraphics[height=0.93cm,loop,autoplay]{8}{figs/result_kth_gif/predrnnpp/pred_152000_3312_}{0}{19} & 
      \includegraphics[height=0.93cm]{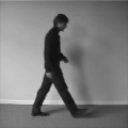} &
      \includegraphics[height=0.93cm]{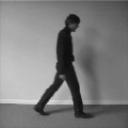} &
      \includegraphics[height=0.93cm]{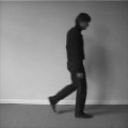} &
      \includegraphics[height=0.93cm]{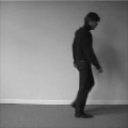}
      \\ [-0.2em]
    \multicolumn{2}{c}{\raisebox{0.3cm}{\scriptsize PredRNN++}} &
    \animategraphics[height=0.93cm,loop,autoplay]{7}{figs/result_mnist_gif/predrnnpp/pred_160000_1456_}{0}{29} & 
    \includegraphics[height=0.93cm]{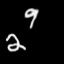} &
    \includegraphics[height=0.93cm]{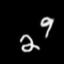} &
    \includegraphics[height=0.93cm]{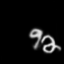} &
    \includegraphics[height=0.93cm]{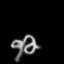} &
    \multicolumn{2}{c}{\raisebox{0.3cm}{\scriptsize PredRNN++}} &
    \animategraphics[height=0.93cm,loop,autoplay]{8}{figs/result_kth_gif/predrnnpp/pred_152000_3312_}{0}{19} & 
    \includegraphics[height=0.93cm]{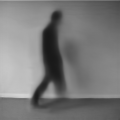} &
    \includegraphics[height=0.93cm]{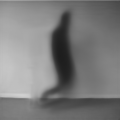} &
    \includegraphics[height=0.93cm]{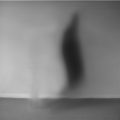} &
    \includegraphics[height=0.93cm]{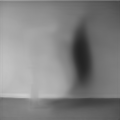}
      \\ [-0.2em]
    \multicolumn{2}{c}{\raisebox{0.3cm}{\scriptsize E3D-LSTM}} &
    \animategraphics[height=0.93cm,loop,autoplay]{7}{figs/result_mnist_gif/e3d/gt_80000_1456_}{0}{29} &
    \includegraphics[height=0.93cm]{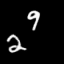} &
    \includegraphics[height=0.93cm]{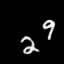} &
    \includegraphics[height=0.93cm]{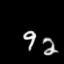} &
    \includegraphics[height=0.93cm]{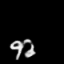} &
    \multicolumn{2}{c}{\raisebox{0.3cm}{\scriptsize E3D-LSTM}} &
    \animategraphics[height=0.93cm,loop,autoplay]{8}{figs/result_kth_gif/e3d/pred_200000_3312_}{0}{19} &
    \includegraphics[height=0.93cm]{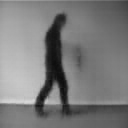} &
    \includegraphics[height=0.93cm]{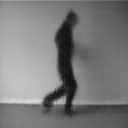} &
    \includegraphics[height=0.93cm]{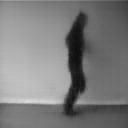} &
    \includegraphics[height=0.94cm]{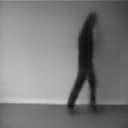} 
    \\ [-0.2em]
    \multicolumn{2}{c}{\raisebox{0.3cm}{\scriptsize \ConvLSTM}} &
    \animategraphics[height=0.93cm,loop,autoplay]{7}{figs/result_mnist_gif/baseline/pred_278_1440_}{0}{29} &
    \includegraphics[height=0.93cm]{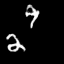} &
    \includegraphics[height=0.93cm]{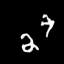} &
    \includegraphics[height=0.93cm]{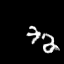} &
    \includegraphics[height=0.93cm]{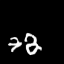} &
    \multicolumn{2}{c}{\raisebox{0.3cm}{\scriptsize \ConvLSTM}} &
    \animategraphics[height=0.93cm,loop,autoplay]{8}{figs/result_kth_gif/baseline/pred_256_3296_}{0}{19} &
    \includegraphics[height=0.93cm]{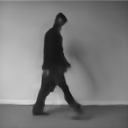} &
    \includegraphics[height=0.93cm]{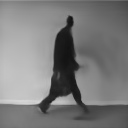} &
    \includegraphics[height=0.93cm]{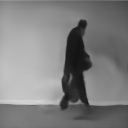} &
    \includegraphics[height=0.93cm]{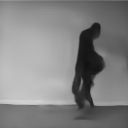} 
    \\ [-0.2em]
    \multicolumn{2}{c}{\raisebox{0.3cm}{\scriptsize \ConvTTLSTM}} &
    \animategraphics[height=0.93cm,loop,autoplay]{7}{figs/result_mnist_gif/ttv4_o3t3r8/pred_297_1440_}{0}{19} &
    \includegraphics[height=0.93cm]{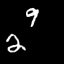} &
    \includegraphics[height=0.93cm]{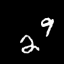} &
    \includegraphics[height=0.93cm]{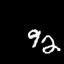} &
    \includegraphics[height=0.93cm]{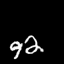} &
    \multicolumn{2}{c}{\raisebox{0.3cm}{\scriptsize \ConvTTLSTM}} &
    \animategraphics[height=0.93cm,loop,autoplay]{8}{figs/result_kth_gif/ttv4_o3t3r8/pred_190_3296_}{0}{19} &
    \includegraphics[height=0.93cm]{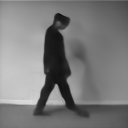} &
    \includegraphics[height=0.93cm]{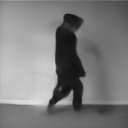} &
    \includegraphics[height=0.93cm]{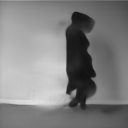} &
    \includegraphics[height=0.93cm]{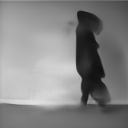} 
  \end{tabular}
  \caption{{\bf 30 frames prediction on Moving-MNIST (left)}, and {\bf 20 frame prediction on KTH action datasets (right)} given 10 input frames. The first frames ($t=1, 11$) are animations. Adobe reader is required to view the animation. Our method generates both semantically plausible and visually crisp images, compared to other approaches.}
  \label{fig:visual}
  \vspace{-0.2cm}
\end{figure*}

\textbf{Multi-frame Video prediction: KTH action dataset. }
First, we test our model on human actions videos.
In Table~\ref{tab:eval}, we report the evaluation on both 20 and 40 frames prediction. Figure~\ref{fig:ssimvslpips} (right) shows the model comparisons with SSIM vs LPIPS and the model size. 
\textbf{(1)} Our model is consistently better than the {\ConvLSTM} baseline for both 20 and 40 frames prediction.
\textbf{(2)} While our proposed {\ConvTTLSTM}s achieve lower SSIM value compared to the state-of-the-art models in 20 frames prediction, they outperform all previous models in LPIPS for both 20 and 40 frames prediction.
\autoref{fig:visual} (right) shows a visual comparison of our model, ConvLSTM baseline, PredRNN++~\cite{wang2018predrnnpp}, and E3D-LSTM~\cite{wang2018eidetic}.
More examples of visual results are presented in \autoref{app:results}.
Overall, our model produces sharper frames and better preserves the details of the human silhouettes, although there exist slight artifacts over time (shifting).
We believe this artifact can be resolved by adding a different loss or an additional technique that help per-pixel motion prediction. 

\textbf{Early activity recognition: Something-Something V2 dataset. }
To demonstrate that our {\ConvTTLSTM}-based prediction model can learn effective representations from videos, we evaluate the models on early activity recognition on the Something-Something V2 dataset. In this task, a model only observes a small fraction (\(25\% - 50\%\)) of frames, and learns to predict future frames. Based on the learned representations of the beginning frames, the model predicts the overall activity of the full video. Intuitively, the learned representation encodes the future information for frame prediction, and the better the representations quality, the higher the classification accuracy. 
As shown in \autoref{tab:eval-s2s-peractivity} and \autoref{tab:eval-s2s} our {\ConvTTLSTM} model consistently outperforms the baseline {\ConvLSTM} and 3D-CNN models as well as E3D-LSTM~\cite{wang2018eidetic} under different ratio of input frames. 
Our experimental setup and architecture follow~\cite{wang2018eidetic}.

\begin{figure}[!htbp]
 \vspace{-0.2cm}
  \centering
  \scriptsize
  \addtolength{\tabcolsep}{-5pt}
  \begin{tabular}{lccccccccccc}
        & \scriptsize Front 25\%& & \scriptsize Front 50\% & & \scriptsize 100\% & \scriptsize Front 25\%& & \scriptsize Front 50\% & & \scriptsize 100\% \\
       & \includegraphics[trim={16.cm 0 0.cm 0.cm},clip, height=1.2cm,width=1.4cm]{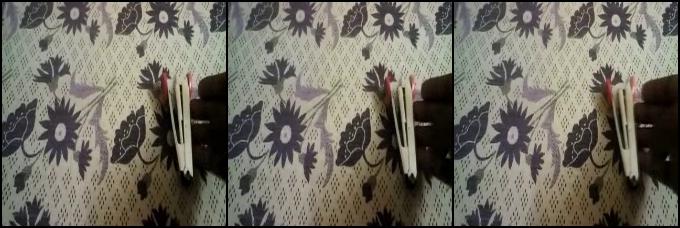} 
      & {\bf ...} & \includegraphics[trim={16.cm 0 0.cm 0.cm},clip, height=1.2cm,width=1.4cm]{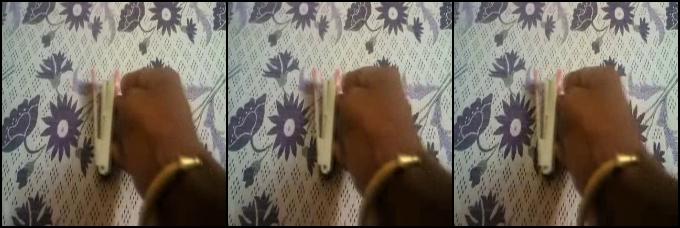}
      & {\bf ....} & \includegraphics[trim={16.cm 0 0.cm 0.cm},clip, height=1.2cm,width=1.4cm]{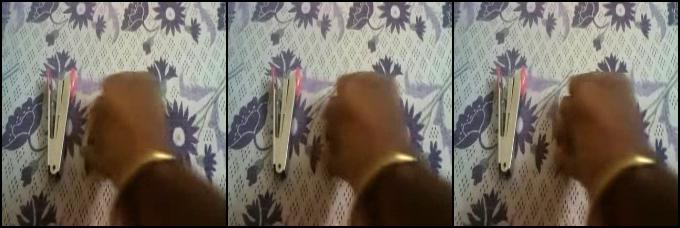} 
      & \includegraphics[trim={16.cm 0 0.cm 0.cm},clip, height=1.2cm,width=1.4cm]{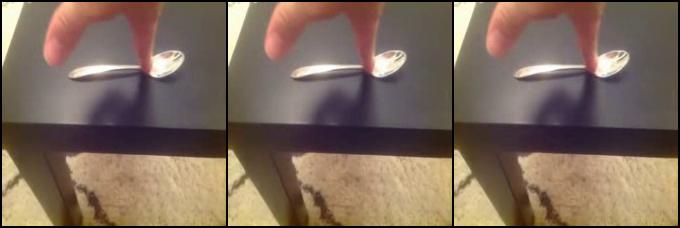} 
      & {\bf ...} & \includegraphics[trim={16.cm 0 0.cm 0.cm},clip, height=1.2cm,width=1.4cm]{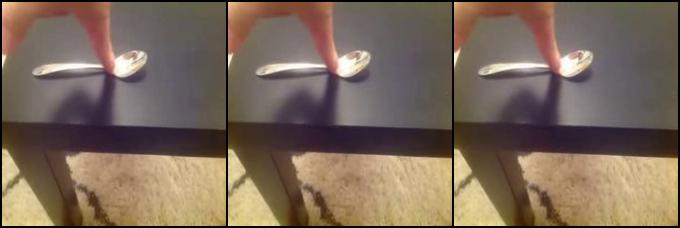}
      & {\bf ....} & \includegraphics[trim={16.cm 0 0.cm 0.cm},clip, height=1.2cm,width=1.4cm]{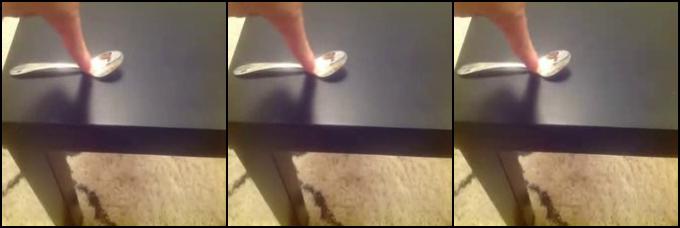} \\ 
       3D-CNN & Wrong (100\%) & & Wrong (100\%) & 
       &\multirow{3}{*}{\thead{\scriptsize Pushing [something] \\ \scriptsize from right to left } } 
        & Wrong (67\%) & & Wrong (100\%) & 
       & \multirow{3}{*}{\thead{\scriptsize Pulling [something] \\ \scriptsize from right to left }} \\ 
       ConvLSTM&  Wrong (84\%) & & Wrong (46\%) & & 
       & Wrong (84\%) & & {\bf Correct (61\%)} & &  \\
       Conv-TT-LSTM&  Wrong (18\%) & & {\bf Correct (100\%)} & & 
       & Wrong (18\%) & &{\bf Correct (98\%)}  & &  \\ [0.5em]
  \end{tabular}
  \caption{{\bf Examples of Early Activity Recognition Results} given 25\% and 50\% of frames on the Something-Something V2 dataset, and ($\cdot$) is the confidence for Correct/Wrong prediction.} 
  \label{fig:visual-s2s}
\vspace{-0.2cm}
\end{figure}

\begin{table}
\setlength{\tabcolsep}{3pt}
\centering
\resizebox{1\textwidth}{!}{
    \caption{\textbf{Per-activity accuracy of early activity recognition on the Something-Something V2 dataset.} We used 41 categories for training. For per-activity evaluation, the 41 categories are grouped into 10 similar activities. The activity mapping are described in \cite{goyal2017something}. Our model substantially outperforms 3D-CNN and ConvLSTM on long-term dynamics such as Moving or Tearing, while achieves marginal improvement on static activities such as Holding or Pouring.}
     \label{tab:eval-s2s-peractivity}
    \begin{tabular}{l l c c c c c c c c c c}
    \toprule
    \bf Model &   \bf \small Input & \bf \small Dropping & \bf \small Holding & \bf \small Moving{LR} & \bf \small Moving{RL} & \bf \small Picking & \bf \small Poking & \bf \small Pouring & \bf \small Putting & \bf \small Showing & \bf \small Tearing \\
    \midrule 
    3D-CNN & \multirow{3}{*}{25\%} & 8.5 & 4.7 & 25.8 & 32.6 & 7.5 & 2.9 & 1.9 & 10.3 & 14.0 & 14.5  \\ 
    ConvLSTM & & 8.5 & {\bf7.0} & 27.4 & 38.8 & 16.8 & 5.9 & 1.9 & 12.0 & 7.0 & 21.2  \\
    {\bf Conv-TT-LSTM} & & {\bf11.5} & 4.7 & {\bf33.9} & {\bf40.8} & {\bf16.8} & {\bf5.9} & {\bf5.7} & {\bf13.6} & {\bf20.9} & {\bf26.0} \\
    \midrule 
    3D-CNN & \multirow{3}{*}{50\%} & 14.6 & 11.6 & 45.2 & 57.1 & 16.8 & 8.8 & 11.3 & 17.4 & 16.3 & 26.0  \\ 
    ConvLSTM & & 21.5 & 7.0 & 43.5 & 47.0 & 15.9 & {\bf14.7} & 5.7 & 20.7 & 16.3 & 30.8  \\
    {\bf Conv-TT-LSTM } & & {\bf24.6} & {\bf11.6} & {\bf56.5} & {\bf57.1} & {\bf27.6} & 5.9 & {\bf13.2} & {\bf25.5} & {\bf37.2} & {\bf46.2} \\
    \bottomrule
  \end{tabular}
}
\vspace{-0.2cm}
\end{table}

\textbf{Multi-frame video prediction: Moving-MNIST-2 dataset. }
We additionally evaluate our model on the Moving-MNIST-2 dataset and show that our model can predict the digits almost perfectly in terms of structure and motion (See ~\autoref{fig:visual}).
Table~\ref{tab:eval} reports the average statistics for 10 and 30 frames prediction, and Figure~\ref{fig:ssimvslpips} (left) shows the comparisons of SSIM vs LPIPS and the model size.
Our {\ConvTTLSTM} models \textbf{(1)} consistently outperform the {\ConvLSTM} baseline for both 10 and 30 frames prediction {\em with fewer parameters};
\textbf{(2)} outperform previous approaches in terms of SSIM and LPIPS (especially on 30 frames prediction), {\em with less than one fifth of the model parameters}.

We reproduce the PredRNN++~\cite{wang2018predrnnpp} and E3D-LSTM~\cite{wang2018eidetic} from the source code~\cite{predrnnpp-source,e3dlstm-source}. 
We find that
\textbf{(1)} PredRNN++ and E3D-LSTM output vague and blurry digits in long-term prediction (especially after 20 steps);  
\textbf{(2)} our {\ConvTTLSTM} produces sharp and realistic digits over all steps. An example of visual comparison is shown in Figure~\ref{fig:visual}, and more visualizations can be found in \autoref{app:results}.

\begin{table}
\vspace{-0.1cm}
\begin{floatrow}
\capbtabbox{%
\setlength{\tabcolsep}{3pt}
  \small{
    \begin{tabular}{l@{\hskip -0.15cm} c c}
    \toprule
    \multirow{2}{*}{\bf Model} & \multicolumn{2}{c}{\bf Input Ratio} \\
    \cmidrule{2-3}
    & \bf Front 25\% & \bf Front 50\% \\
    \midrule 
    3D-CNN* & 9.11 & 10.30 \\
    E3D-LSTM*~\cite{wang2018eidetic} & 14.59 & 22.73 \\
    \midrule
    3D-CNN & 13.26 & 20.72 \\
    ConvLSTM & 15.46 & 21.97 \\
    Conv-TT-LSTM (ours) & {\bf 19.53} & {\bf 30.05} \\
    \bottomrule
  \end{tabular}
  }
}{%
\vspace{0.18cm}
 \caption{\textbf{Early activity recognition on the Something-Something V2 dataset} using 41 categories as \cite{wang2018eidetic}. (*) indicates the result by~\cite{wang2018eidetic}.}
 \label{tab:eval-s2s}
}
\capbtabbox{%
\setlength{\tabcolsep}{3pt}
\centering
\small{
    \begin{tabular}{l c c c}
    \toprule
     & \bf \footnotesize MSE($\times 10^{-3}$) & \bf \footnotesize SSIM & \bf \footnotesize LPIPS \\
    \midrule
    \multicolumn{4}{c}{\footnotesize {\CTTD} with $1 \times 1$ filters (similar to standard {\TTD})} \\
    \midrule
    single order & 31.52 & 0.810 & 148.7 \\
    order 3 & 34.84 & 0.800 & 151.2 \\
    \midrule
    \multicolumn{4}{c}{\footnotesize {\CTTD} with $5 \times 5$ filters} \\
    \midrule
    single order & 33.08 & 0.806 & 140.1  \\
    order 3 & {\bf 28.88} & {\bf 0.831} & {\bf 104.1} \\
    \bottomrule
  \end{tabular}
  }
}{%
  \caption{\textbf{Ablation studies of higher-order {\ConvTTLSTM} on Moving-MNIST-2 dataset}. The models are tested for \(10\) to \(30\) frames prediction. 
  }
\label{tab:eval-ablation-conv}
}
\end{floatrow}
\vspace{-0.6cm}
\end{table}

\begin{table*}
\setlength{\tabcolsep}{4pt}
\centering
\caption{{\bf Evaluation of multi-steps prediction on the KTH action (top) and Moving-MNIST-2 (bottom) datasets.} Higher PSNR/SSIM and lower MSE/LPIPS values indicate better predictive results. \# of FLOPs denotes the multiplications for one-step prediction per sample, and Time(m) represents the clock time (in minutes) required by training the model for one epoch (10,000 samples)}
\label{tab:eval}
\resizebox{1\textwidth}{!}{
    \setlength{\tabcolsep}{1.1pt}
    \begin{tabular}{l l  c c c  c c c  c c c}
    \toprule
     & \multicolumn{1}{c}{\multirow{2}{*}{\bf Method}} & \multicolumn{3}{c}{\bf (10 -> 20)} & \multicolumn{3}{c}{\bf (10 -> 40)} & \multicolumn{3}{c}{\bf Complexities} \\    
     \cmidrule(l{3pt}r{3pt}){3-5}  \cmidrule(l{3pt}r{3pt}){6-8} \cmidrule(l{3pt}r{3pt}){9-11}
    & & \bf PSNR & \bf SSIM & \bf LPIPS & \bf PSNR & \bf SSIM & \bf LPIPS & \bf \# Params. & \bf \# FLOPS & \bf Time(m) \\
    \midrule
    \parbox[t]{5mm}{
        \multirow{7}{*}{\rotatebox[origin=c]{90}{\textbf{KTH action}}}
    }
    & ConvLSTM~\cite{xingjian2015convolutional} & 23.58 & 0.712 & - & 22.85 & 0.639 & - & 7.58M & - & - \\
    & MCNET~\cite{villegas2017decomposing} & 25.95 & 0.804 & - & - & - & - & - & - & - \\
    & PredRNN++~\cite{wang2018predrnnpp}~(retrained~\cite{predrnnpp-source}) & 28.62 & 0.888 & 228.9 & 26.94 & 0.865 & 279.0 & 15.05M & - & - \\
    & E3D-LSTM~\cite{wang2018eidetic}~(retrained~\cite{e3dlstm-source}) & 27.92 & 0.893 & 298.4 & 26.55 & 0.878 & 328.8 & 41.94M & - & - \\ 
    \cmidrule[\heavyrulewidth]{2-11}
    & ConvLSTM (baseline) & 28.21 & 0.903 & 137.1 & 26.01 & 0.876 & 201.3 & 3.97M & 55.83G & 26.2 \\ 
  & {\ConvLSTM} (classic TTD~\citep{yang2017tensor, garipov2016ultimate}) & 27.70 & 0.897 & 141.5 & 25.89 & 0.872 &  191.7 & 2.21M & - & - \\
    \cmidrule{2-11}
    & \textbf{Conv-TT-LSTM (Ours)}  & 28.36 & {0.907} & {133.4} & 26.11 & {0.882} & 191.2 & 2.69M & 37.83G & 27.3 \\
    \bottomrule
\end{tabular}
}%
\newline
\resizebox{1\textwidth}{!}{
    \setlength{\tabcolsep}{1.2pt}
    \begin{tabular}{l l  c c c  c c c  c c c}
    \toprule
    & \multicolumn{1}{c}{\multirow{2}{*}{\bf Method}} & \multicolumn{3}{c}{\bf (10 -> 10)} & \multicolumn{3}{c}{\bf (10 -> 30)} & \multicolumn{2}{c}{\bf Complexities} \\
     \cmidrule(l{3pt}r{3pt}){3-5}  \cmidrule(l{3pt}r{3pt}){6-8} \cmidrule(l{3pt}r{3pt}){9-11}
    & & \bf MSE & \bf SSIM & \bf LPIPS & \bf MSE & \bf SSIM & \bf LPIPS & \bf  \# Params. & \bf \# FLOPS & \bf Time(m) \\
    \midrule
    \parbox[t]{5mm}{
        \multirow{7}{*}{\rotatebox[origin=c]{90}{\textbf{Moving-MNIST}}}
    }
    & ConvLSTM~\cite{xingjian2015convolutional} 
    & 25.22 & 0.713 & - & 38.13 & 0.595 & - & 7.58M & - & - \\
    & VPN~\cite{kalchbrenner2017video} 
    & 15.65 & 0.870 & - & 31.64 & 0.620 & - & - & - & - \\
    & PredRNN++~\cite{wang2018predrnnpp}~(retrained~\cite{predrnnpp-source}) & 10.29 & 0.913 & 59.51 & \text{20.53} & 0.834 & 139.9 & 15.05M & - & - \\ 
    & E3D-LSTM~\cite{wang2018eidetic}~(pretrained~\cite{e3dlstm-source}) & 20.23 & 0.869 & 76.12 & 32.37 & 0.803 & 150.3 & 41.94M & - & - \\ 
    \cmidrule[\heavyrulewidth]{2-11}
    & {\ConvLSTM} (baseline) & 18.17 & 0.882 & 67.13 & 33.08 & 0.806 & 140.1 & 3.97M & 15.88G & 6.35 \\ 
    & {\ConvLSTM} (classic TTD~\citep{yang2017tensor, garipov2016ultimate})  & 16.78 & 0.890 & 57.90 & 29.07 & 0.815 & 126.4 & 2.20M & - & - \\
    \cmidrule{2-11}
   & \textbf{{\ConvTTLSTM} (Ours)} & 12.96 & \textbf{0.915} & \textbf{40.54} & 25.81 & \textbf{0.840} & \textbf{90.38} & 2.69M & 10.76G & 7.40 \\ 
    \bottomrule
  \end{tabular}
}%
\end{table*}

\section{Discussion}\vspace{-5pt}
\label{sec:discussion}
In this section, we further justify the importance of the proposed modules, {\em {\CTTDlong}} (\CTTD) and the {\em preprocessing module}. 
We also explain the computational complexity of our model, and the difficulties of spatio-temporal learning with Transformer~\citep{vaswani2017attention}. 

\textbf{Importance of encoding higher-order correlations in a convolutional manner.}
Two key differences between {\CTTD} and existing low-rank decompositions are {\em higher-order decomposition} and {\em convolutional operations}.
To verify their impact,
we compare the performance of two ablated models against our {\CTTD}-base model in Table~\ref{tab:eval-ablation-conv}. 
The single order means that the higher-order model is replaced with a first-order model (\text{tensor order} = $1$). By replacing $5 \times 5$ filters to $1 \times 1$, the convolutions are removed, and the {\CTTD} reduces to a standard {\TTDlong}.
The results show a decrease in performance: the ablated models at best achieve similar performance of {\ConvLSTM} baseline, demonstrating that both higher-order model and convolutional operations are necessary. 

\textbf{Importance of the preprocessing module.} 
There could be other ways to incorporate previous hidden states into the \(\mathsf{CTT}\) module.
One is to reduce the number of channels while keeping the number of steps; the other is to reuse the concatenation of all previous states for each input to \(\mathsf{CTT}\). The former fails due to gradient vanishing/exploding problem, while the latter has a tube-shaped receptive field that fails to distinguish more recent steps and the ones from the remote history. 

\textbf{Computational complexity.} 
The number of FLOPS for all models are compared in~\autoref{tab:eval}. 
Our {\ConvTTLSTM} model has a lower computational complexity and fewer parameters than other models under comparison.
This efficiency is made possible by a linear algorithm for the convolutional tensor-train module in Eq~\eqref{eq:conv-tt-lstm-v1}, which is derived in \autoref{app:tensor-trains}. 

\textbf{Trade-off between FLOPs and latency.}
Notice that a lower number of FLOPS does not necessarily lead to faster computation due to the sequential nature of convolutional tensor-train module.
In \autoref{app:tensor-trains}, we introduce two algorithms. While {Alg.~\ref{alg:conv-tt-lstm-fast} significantly decreases the complexity in FLOPs}, it also lowers the degree of parallelism. However, {Alg.~\ref{alg:conv-tt-lstm-slow} shows how our model can be parallelized}. Ideally, these two algorithms can be combined using CUDA multi-streams (execute multiple kernels in parallel): use Alg.~\ref{alg:conv-tt-lstm-slow} for the beginning iterations of $i$ and Alg.~\ref{alg:conv-tt-lstm-fast} for the later ones (the beginning ones have smaller kernel sizes).
In our current implementation, we use Alg.\ 2 to reduce the GPU memory requirement, and the run-time is 27.3 mins (37.83 GFLOPs) for Conv-TT-LSTM verse 26.2 mins (55.83 GFLOPs) for ConvLSTM (per epoch on KTH), as shown in~\autoref{tab:eval}.

\textbf{Classic Tensor-Train Decomposition for RNN compression.}
Classic {\em \TTDlong} (\TTD)\citep{oseledets2011tensor} is traditionally used to compress fully-connected and convolutional layer in a feed-forward network~\citep{novikov2015tensorizing, garipov2016ultimate}, 
where the parameters in each layer are reshaped into a higher-order tensor 
and stored in a factorized form.
\citet{yang2017tensor} applies this idea to RNNs, 
and proposes to compress the parameters in the input-hidden transition matrix~\citep{novikov2015tensorizing}.

There are three major differences between our work and \citet{yang2017tensor}: 
\begin{itemize}[leftmargin=*, itemsep=0pt, topsep=0pt]
    \item {Single-order LSTM v.s.\ Higher-order \ConvLSTM.} 
\citet{yang2017tensor} consider a first-order fully-connected LSTM~\citep{hochreiter1997long} for compression, while our method aims to compress a higher-order \ConvLSTMlong model.
    \item {Classic decomposition v.s.\ Convolutional decomposition.} 
\citet{yang2017tensor} relies on the classic {\TTD}, while our proposed {\em \CTTDlong} (\CTTD) factorizes the tensor with convolutions in addition to inner products; our decomposition is designed to preserve spatial structures in spatio-temporal data.
    \item {Compression of input-hidden matrix v.s.\ hidden-to-hidden convolutional kernels.}
\citet{yang2017tensor} only
compresses input-hidden transition \(\mymatrix{W}\) in LSTM, but
our {\CTTD} compresses a sequence of convolutional kernels \(\{\tensorSup{K}{1}, \cdots, \tensorSup{K}{N}\}\) for different time steps simultaneously (see Eq.~\eqref{eq:higher-order-convolutional}).
\end{itemize}

To understand the necessity of our design for long-term spatio-temporal dynamics, we develop a compressed {\ConvLSTM} following the same idea in \citep{novikov2015tensorizing, garipov2016ultimate, yang2017tensor}, which stores the parameters for input-hidden transition \(\mytensor{W}\) in a tensor-train format \(\mytensor{W} = \textsf{TT}(\{\tensorSup{G}{i}\}_{i = 0}^{N - 1})\) (where \(N\) denotes the order of the decomposition, i.e.\ the number of factors):
\begin{equation}
[ \mathcal{I}{(t)}; \mathcal{F}{(t)}; \mathcal{\Tilde{C}}{(t)}; \mathcal{O}{(t)} ]
= \sigma( \textsf{TT}(\{\tensorSup{G}{i}\}_{i = 0}^{N - 1}) \ast \mathcal{X}{(t)} + \mathcal{K} \ast \mathcal{H}{(t-1)})
\end{equation}
Since the transition in {\ConvLSTM} is characterized as a convolutional layer, we follow the approach by \citet{garipov2016ultimate} and represent \(\mytensor{W}\) with size \([K \times K \times C_\text{out} \times C_\text{in}]\) using \(N\) factors: {\bf(1)} The \(4\)-th order tensor \(W\) is reshaped to an \(2M\)-th order tensor \(\widetilde{\mytensor{W}}\) with size \([K \times K \times T_1 \cdots \times T_{N-1} \times S_{1} \cdots \times S_{N-1}]\) and \(C_\text{out} = \prod_{i = 1}^{N-1} T_i\), \(C_\text{in} = \prod_{i = 1}^{N-1} S_i\); {\bf(2)} The tensor \(\widetilde{\mytensor{W}}\) is factorized using \TTD as
\begin{equation}
\widetilde{\mytensor{W}}_{i, j, t_1, \cdots, t_{N-1}, s_1, \cdots, s_{N-1}} \triangleq \sum_{r_0, \cdots, r_{N-1}} \tensorInd{G}{i, j, r_0}{0} \tensorInd{G}{t_1, s_1, r_0, r_1}{1} \cdots \tensorInd{G}{i, j, r_{N-1}}{N-1}, 
\end{equation}
where \(\tensorSup{G}{0}\) has size \([K \times K \times R_0]\),
\(\tensorSup{G}{i}\) has  \( [T_{i} \times S_{i} \times R_{i-1} \times R_{i}]\) 
for \(0 < i < N-1\),
and \(\tensorSup{G}{N-1}\) has \([T_{N-1} \times S_{N-1} \times R_{N-1}]\).
A comparison against the uncompressed {\ConvLSTM} and our {\ConvTTLSTM} is presented in Table~\ref{tab:eval}. We observe that our model outperforms this method on MNIST and KTH (except LPIPS on KTH) with similar number of parameters.

\textbf{Transformer for spatio-temporal learning.}
Transformer~\citep{vaswani2017attention} is a popular predictive model based on attention mechanism, which is very successful in natural language processing~\citep{devlin2019bert}.
However, Transformer has prohibitive limitations when it comes to video understanding, due to excessive needs for both memory and computation.
While language modeling only involves temporal attention,
video understanding requires attention on spatial dimensions as well~\citep{weissenborn2019scaling}.
Moreover, since attention mechanism is not designed to preserve the spatial structures, Transformer additionally requires auxiliary components including autoregressive module and multi-resolution upscaling when applied on spatial data~\citep{parmar2018image, menick2018generating, weissenborn2019scaling}.
Our {\ConvTTLSTM} incorporates a large spatio-temporal context, but with a compact, efficient and structure-preserving operator without additional components. 
\section{Related Work}
\label{sec:related}

{\bf Tensor decompositions.}
Tensor decompositions such as CP, Tucker or Tensor-Train~\citep{kolda2009tensor,oseledets2011tensor}, 
are widely used for dimensionality reduction~\citep{cichocki2016tensor} and learning probabilistic models~\citep{anandkumar2014tensor}.
These tensor factorization techniques have also been widely used in deep learning, to improve performance, speed-up computation and compress the deep neural networks~\citep{lebedev2014speeding, kim2015compression,novikov2015tensorizing, kossaifi2017tensor,yang2016deep,su2018tensorized, kolbeinsson2019stochastically,kossaifi2019efficient}, recurrent networks~\citep{tjandra2017compressing, yang2017tensor} and Transformers~\citep{ma2019tensorized}.
\citet{yang2017tensor} has proposed tensor-train RNNs to compress both inputs-states and states-states matrices within each cell with {\TTD} by reshaping the matrices into tensors, and showed improvement for video classification.

Departing from prior works that rely on existing, well-established tensor decompositions,
here we propose a novel {\em \CTTDlong} (\CTTD) 
that is designed to enable efficient and compact higher-order convolutional recurrent networks.
Unlike~\citet{yang2017tensor}, we aim to compress higher-order {\ConvLSTM}, rather than first-order fully-connected LSTM. We further propose Convolutional Tensor-Train decomposition to preserve spatial structure after compression.

{\bf Spatio-temporal prediction models.}
Prior prediction models have focused on predicting short-term video~\citep{lotter2016deep,byeon2018contextvp} or decomposing motion and contents~\citep{finn2016unsupervised,villegas2017decomposing,denton2017unsupervised,hsieh2018learning}.
Many of these works use {\ConvLSTM} as a base module, which deploys 2D convolutional operations in LSTM to efficiently exploit spatio-temporal information. 
Some works modified the standard {\ConvLSTM} to better capture spatio-temporal correlations \citep{wang2017predrnn, wang2018predrnnpp}.
\citet{byeon2018contextvp}
demonstrated strong performance using a deep {\ConvLSTM} network as a baseline, which is used as the base architecture in the present paper.
\section{Conclusion}
In this paper, we proposed a fully-convolutional higher-order LSTM model for spatio-temporal data. 
To make the approach computationally and memory feasible, we proposed a novel convolutional tensor-train decomposition that jointly parameterizes the convolutions and naturally encodes temporal dependencies. 
The result is a compact model that outperforms prior work on video prediction, including something-something V2, moving-MNIST-2 and the KTH action datasets.
Going forward, we plan to investigate our CTT module in a framework that spans not only higher-order RNNs but also Transformer-like architectures for efficient spatio-temporal learning.

\bibliography{\includehome/supp_bib}
\bibliographystyle{unsrtnat}
\clearpage

\appendix
\begin{center}
{\large\bf Appendix: Convolutional Tensor-Train LSTM for Spatio-temporal Learning}
\end{center}

In the supplementary material, 
we first provide a constructive proof that our approach can be computed in linear time. 
We then provide thorough implementation details for all experiments, and perform extra ablation studies of our model, demonstrating that our {\ConvTTLSTM} model is general and outperforms regular {\ConvLSTM} regardless of the architecture or setting used. 
Finally, we provide additional visualizations of our experimental results.

To facilitate the reading of our paper, we provide a Table of notation in \autoref{tab:notation}.

\begin{table}[ht]
    \centering
    \begin{tabular}[t]{c c c}
        \toprule
        \textbf{Symbol} & \textbf{Meaning} & \textbf{Value} or \textbf{Size} \\
        \midrule
        $H$ 
            & Height of feature map 
            & \multirow{4}{*}{-}
        \\
        $W$ & Width of feature map
            & 
        \\
        $C_{\text{in}}$
            & \# of input channels
            & 
        \\
        $C_{\text{out}}$ 
            & \# of output channels
            & 
        \\
        \(t\) 
            & Current time step
            & -
        \\
        \(\mytensor{W}\)
            & Weights for \(\tensorSup{X}{t}\)  
            & \([K \times K \times 4C_{\text{out}} \times C_{\text{in}}]\)
        \\
        \(\tensorSup{X}{t}\) 
            & Input features
            & \([H \times W \times C_{\text{in}}]\)
        \\ \midrule
        \( \tensorSup{H}{t}\)
            & Hidden state
            & \multirow{6}{*}{\([H \times W \times C_{\text{out}}]\)}
        \\
        \(\tensorSup{C}{t}\)
            & Cell state
            & 
        \\
        \(\tensorSup{I}{t}\) 
            & Input gate
            &  
        \\
        \(\tensorSup{F}{t}\)
            & Forget gate
            &  
        \\
        \(\tensorSup{\tilde{C}}{t}\)
            & Cell memory
            & 
        \\
        \(\tensorSup{O}{t}\)
            & Output gate
            & 
        \\ \midrule
        $\Phi$
            & Mapping function for higher-order {\RNN}
            & -
        \\
        \(M\)
            & order of higher-order {\RNN}
            & \multirow{2}{*}{\(M \geq N\)}
        \\
        \(N\)
            & Order of {\CTTD}
            & 
        \\
        \(K\)
            & Initial filter size
            & \multirow{2}{*}{\(\scalarSup{K}{0} = K \)}
        \\ 
        \(\scalarSup{K}{i}\)
            & Filter size in \(\tensorSup{\Tilde{K}}{i}\)
            &
        \\
        \(\scalarSup{C}{i}\) 
            & \# channels in \(\tensorSup{\Tilde{H}}{i}\)
            & \(\scalarSup{C}{0} = 4 C_{\text{out}} \)
        \\ 
        $\tensorSup{G}{i}$
            & Factors in the {\CTTD} 
            & \([\scalarSup{K}{0} \times \scalarSup{K}{0} \times \scalarSup{C}{i} \times \scalarSup{C}{i-1}]\)
        \\ \midrule
        \(D\)
            & Size of sliding window 
            & \(D =  M - N +1\)
        \\
        \(\tensorSup{P}{i}\)
            & Preprocessing kernel 
            & \([D \times K \times K \times C_{\text{out}} \times \scalarSup{C}{i}]\)
        \\
        \(\tensorSup{\Tilde{H}}{i}\)
            & Pre-processed hidden state
            & \([H \times W \times \scalarSup{C}{i}]\)
        \\
        \(\tensorSup{K}{i}\) 
            & Weights for \(\tensorSup{\Tilde{H}}{i}\)
            & \([\scalarSup{K}{i} \times \scalarSup{K}{i} \times \scalarSup{C}{i} \times \scalarSup{C}{0}] \) 
        \\
        \bottomrule
    \end{tabular}
    \caption{\textbf{Table of notations.}}
    \label{tab:notation}
\end{table}

\section{An Efficient Algorithm for Convolutional Tensor-Train Module}
\label{app:tensor-trains}

In this section, we prove that our convolutional tensor-train module, \(\mathsf{CTT}\) (Eq.7 in main paper), can be evaluated with linear computational complexity.
Our proof is constructive and readily provides an algorithm for computing \(\mathsf{CTT}\) in linear time.

First, let's recall the formulation of the \(\mathsf{CTT}\) function:
{\small
\begin{equation}
\Phi = \mathsf{CTT} ( \tensorSup{H}{t-1}, \cdots, \tensorSup{H}{t-N}; ~\tensorSup{G}{1}, \cdots, \tensorSup{G}{N} ) = \sum_{i = 1}^{N} \tensorSup{K}{i} \ast \tensorSup{H}{t-i}.
\label{eq:CTT}
\end{equation}}%
where each kernel \(\tensorSup{K}{i}\) is factorized by {\em \CTTDlong} (\CTTD): 
{\small
\begin{equation}
\tensorInd{K}{\bm{:}, \bm{:}, c_{i}, c_0}{i} \triangleq \mathsf{CTTD}\left(\{\tensorSup{G}{j}\}_{j = 1}^{i} \right) 
= \sum_{c_{i - 1} = 1}^{\scalarSup{C}{i-1}} \cdots \sum_{c_1 = 1}^{\scalarSup{C}{1}} \tensorInd{G}{\bm{:}, \bm{:}, c_{i}, c_{i-1}}{i} \ast \cdots \ast \tensorInd{G}{\bm{:}, \bm{:}, c_1, c_0}{1}, \forall i \in [N]
\end{equation}}%
However, a naive algorithm that first reconstruct all the kernels \(\tensorSup{K}{i}\), then applies Eq.\eqref{eq:CTT} results in a computational complexity of \(O(N^3)\), as illustrated in \autoref{alg:conv-tt-lstm-slow}. 
To scale our approach to higher-order models (i.e.\ larger \(N\)), we need a more efficient implementation of the function \(\mathsf{CTT}\).

{\small
\begin{algorithm}[!htbp]
\DontPrintSemicolon
\KwIn{current input \(\tensorSup{X}{t}\), previous cell state \(\tensorSup{C}{t-1}\),} 
\myinput{\(M\) previous hidden states \(\{\tensorSup{H}{t-1}, \cdots, \tensorSup{H}{t-M}\}\)} 
{\bf Output: } new hidden state \(\tensorSup{H}{t}\), new cell state \(\tensorSup{C}{t}\) \;
{\bf Initialization: }\(\tensorSup{K}{0} = 1 \); ~ \(\mytensor{V} = 0 \) \;
\tcc{Convolutional Tensor-Train (CTT) module}
\For{\(i = 1\) \KwTo \(N\)}{ 
    \tcc{preprocessing module}
    \tcp{compress the states from a sliding window} 
    \(\tensorSup{\Tilde{H}}{i} = \tensorSup{P}{i} \ast \left[\tensorSup{H}{t-i}; \cdots; \tensorSup{H}{t-i+N-M}\right] \)  \;
    \tcp{recursively construct the kernel}
    \(\tensorSup{K}{i} = \tensorSup{G}{i} \ast \tensorSup{K}{i-1}\) \;
    \tcp{accumulate the output}
    \(\mytensor{V} = \mytensor{V} + \tensorSup{K}{i} \ast \tensorSup{\Tilde{H}}{i}\) 
}
\tcc{Long-Short Term Memory (LSTM)}
\( \left[\tensorSup{I}{t}; \tensorSup{F}{t};  
\vectorSup{\tilde{C}}{t}; \tensorSup{O}{t} \right] = \sigma(\mytensor{W} \ast \tensorSup{X}{t} + \mytensor{V} )\) \;
\(\tensorSup{C}{t} = \tensorSup{C}{t-1} + \tensorSup{\Tilde{C}}{t} \circ \tensorSup{I}{t}; ~ \tensorSup{H}{t} = \tensorSup{O}{t} \circ \sigma(\tensorSup{C}{t}) \) \;
return \(\tensorSup{H}{t}\), \(\tensorSup{C}{t}\)\;
\caption{{\bf Convolutional Tensor-Train LSTM} (Original: \(T(N) = O(N^3)\)).}
\label{alg:conv-tt-lstm-slow}
\end{algorithm}}%

\textbf{Recursive evaluation.} We will prove that \(\mathsf{CTT}\) can be evaluated backward recursively using 
{\small
\begin{equation}
\tensorInd{V}{\bm{:}, \bm{:}, c_{i-1}}{i-1} = \sum_{c_i = 1}^{\scalarSup{C}{i}} \tensorInd{G}{\bm{:}, \bm{:}, c_{i}, c_{i-1}}{i} \ast \left(\tensorInd{V}{\bm{:}, \bm{:}, c_{i}}{i} + \tensorInd{H}{\bm{:}, \bm{:}, c_{i}}{i} \right), ~ i = N, N-1, \cdots, 0
\label{eq:CTT-recursive}
\end{equation}
}%
where \(\tensorSup{V}{N}\) is initialized as zeros, and the final output of \(\mathsf{CTT}\) is equal to \(\tensorSup{V}{0}\).

\begin{proof}
First, we note that \(\tensorSup{K}{i}\) can be represented recursively in terms of \(\tensorSup{K}{i-1}\) and \(\tensorSup{G}{i}\):
{\small
\begin{equation}
\tensorInd{K}{\bm{:}, \bm{:}, c_i, c_0}{i} = \sum_{c_{i-1} = 1}^{\scalarSup{C}{i-1}} \tensorInd{G}{\bm{:}, \bm{:}, c_{i}, c_{i-1}}{i} \ast \tensorInd{K}{\bm{:}, \bm{:}, c_{i-1}, c_0}{i-1}
\label{eq:CTTD-recursive}
\end{equation}
}%
with \(\tensorSup{K}{1} = \tensorSup{G}{1}\). Next, we aim to {\bf inductively} prove the following holds for any \(n \in [N]\):  
{\small
\begin{equation}
\Phi_{\bm{:}, \bm{:}, c_0} = \sum_{i = 1}^{n} \sum_{c_i = 1}^{\scalarSup{C}{i}} \tensorInd{K}{\bm{:}, \bm{:}, c_i, c_0}{i} \ast \tensorInd{H}{\bm{:}, \bm{:}, c_i}{t-i} + \sum_{c_n = 1}^{\scalarSup{C}{n}} \tensorInd{K}{\bm{:}, \bm{:}, c_n, c_0}{n} \ast \tensorInd{V}{\bm{:}, \bm{:}, c_{n}}{n},
\end{equation}
}%
and therefore it holds for \(n = 1\),
\(
\Phi_{\bm{:}, \bm{:}, c_0} = \sum_{c_1 = 1}^{\scalarSup{C}{1}} \tensorInd{G}{\bm{:}, \bm{:}, c_1, c_0}{1} \ast (\tensorInd{V}{\bm{:}, \bm{:}, c_1}{1} + \tensorInd{H}{\bm{:}, \bm{:}, c_1}{1}) = \tensorInd{V}{\bm{:}, \bm{:}, c_0}{0}
\).

Notice that the case \(n = N\) is obvious by the definition of \(\mathsf{CTT}\) and the zero initialization of \(\tensorSup{V}{N}\). Therefore the remaining of this proof is to induce the case \(n = N-1\) from \(n = N\).
{\small
\begin{align}
\Phi_{\bm{:}, \bm{:}, c_0} & = \sum_{i = 1}^{N} \sum_{c_i = 1}^{\scalarSup{C}{i}} \tensorInd{K}{\bm{:}, \bm{:}, c_i, c_0}{i} \ast \tensorInd{H}{\bm{:}, \bm{:}, c_i}{t-i} + \sum_{c_N = 1}^{\scalarSup{C}{N}} \tensorInd{K}{\bm{:}, \bm{:}, c_N, c_0}{N} \ast \tensorInd{V}{\bm{:}, \bm{:}, c_{N}}{N} \\
& = \sum_{i = 1}^{N - 1} \sum_{c_i = 1}^{\scalarSup{C}{i}} \tensorInd{K}{\bm{:}, \bm{:}, c_i, c_0}{i} \ast \tensorInd{H}{\bm{:}, \bm{:}, c_i}{t-i} + 
\underbrace{\sum_{c_N = 1}^{\scalarSup{C}{N}} \tensorInd{K}{\bm{:}, \bm{:}, c_N, c_0}{N} \ast \left(\tensorInd{H}{\bm{:}, \bm{:}, c_N}{N} + \tensorInd{H}{\bm{:}, \bm{:}, c_N}{N}\right)}
\end{align}
}%
Notice that the second term can be rearranged as
{\small
\begin{align}
& \sum_{c_{N} = 1}^{\scalarSup{C}{N}} \tensorInd{K}{\bm{:}, \bm{:}, c_N, c_0}{N} \ast \left(\tensorInd{H}{\bm{:}, \bm{:}, c_N}{N} + \tensorInd{H}{\bm{:}, \bm{:}, c_N}{N}\right) \\
= ~ & \sum_{c_{N} = 1}^{\scalarSup{C}{N}} \bigg[ \underbrace{\sum_{c_{N-1} = 1}^{\scalarSup{C}{N-1}} \tensorInd{G}{\bm{:}, \bm{:}, c_N, c_{N-1}}{N-1} \ast \tensorInd{K}{\bm{:}, \bm{:}, c_{N-1}, c_0}{N-1}}_{\tensorInd{K}{\bm{:}, \bm{:}, c_N, c_0}{N}, ~ \text{\scriptsize by Eq.\eqref{eq:CTTD-recursive}}} \bigg] \ast \left(\tensorInd{H}{\bm{:}, \bm{:}, c_N}{N} + \tensorInd{H}{\bm{:}, \bm{:}, c_N}{N}\right) 
\label{eq:proof-key1}
\end{align}}%
{\small
\begin{align}
= ~ & \sum_{c_{N-1} = 1}^{C_{N-1}} \tensorInd{K}{\bm{:}, \bm{:}, c_{N-1}, c_0}{N-1} \ast \bigg[ \underbrace{\sum_{c_{N} = 1}^{\scalarSup{C}{N}} \tensorInd{G}{\bm{:}, \bm{:}, c_N, c_{N-1}}{N-1} \ast \left(\tensorInd{H}{\bm{:}, \bm{:}, c_N}{N} + \tensorInd{H}{\bm{:}, \bm{:}, c_N}{N}\right)}_{\tensorInd{V}{\bm{:}, \bm{:}, N-1}{N-1}, ~ \text{\scriptsize by Eq.\eqref{eq:CTT-recursive}}} \bigg] 
\label{eq:proof-key2} \\
= ~ & \sum_{c_{N-1} = 1}^{\scalarSup{C}{N-1}} \tensorInd{K}{\bm{:}, \bm{:}, c_{N-1}, c_0}{N-1} \ast \tensorInd{V}{\bm{:}, \bm{:}, N-1}{N-1}
\end{align}
}%
where Eq.\eqref{eq:proof-key1} uses the recursive formula in Eq.\eqref{eq:CTTD-recursive}, and Eq.\eqref{eq:proof-key2} is by definition of \(\tensorSup{V}{N-1}\) in Eq.\eqref{eq:CTT-recursive}. Therefore, we show that the case \(n = N - 1\) also holds
{\small
\begin{align}
\Phi_{\bm{:}, \bm{:}, c_0}
& = \sum_{i = 1}^{N-1} \sum_{c_i = 1}^{\scalarSup{C}{i}} \tensorInd{K}{\bm{:}, \bm{:}, c_i, c_0}{i} \ast \tensorInd{H}{\bm{:}, \bm{:}, c_i}{t-i} + \sum_{c_{N-1} = 1}^{\scalarSup{C}{N-1}} \tensorInd{K}{\bm{:}, \bm{:}, c_{N-1}, c_0}{N-1} \ast \tensorInd{V}{\bm{:}, \bm{:}, N-1}{N-1}
\end{align}
}%
which completes the induction from \(n = N\) to \(n = N - 1\). 
\end{proof}

{\small
\begin{algorithm}[!htbp]
\DontPrintSemicolon
\KwIn{current input \(\tensorSup{X}{t}\), previous cell state \(\tensorSup{C}{t-1}\),} 
\myinput{\(M\) previous hidden states \(\{\tensorSup{H}{t-1}, \cdots, \tensorSup{H}{t-M}\}\)} 
{\bf Output: } new hidden state \(\tensorSup{H}{t}\), new cell state \(\tensorSup{C}{t}\) \;
{\bf Initialization: }\(\tensorSup{K}{0} = 1 \); ~ \(\tensorSup{V}{N} = 0 \) \;
\tcc{Convolutional Tensor-Train (CTT) module}
\For{\(i = N\) \KwTo \(1\)}{ 
    \tcc{preprocessing module}
    \tcp{compress the states from a sliding window} 
    \(\tensorSup{\Tilde{H}}{i} = \tensorSup{P}{i} \ast \left[\tensorSup{H}{t-i}; \cdots; \tensorSup{H}{t-i+N-M}\right] \)  \; 
    \tcp{recursively compute the intermediate results}
    \(\tensorSup{V}{i-1} = \tensorSup{G}{i} \ast \left(\tensorSup{V}{i} + \tensorSup{\Tilde{H}}{i} \right) \);
}
\tcc{Long-Short Term Memory (LSTM)}
\(\left[\tensorSup{I}{t}; \tensorSup{F}{t};  
\vectorSup{\tilde{C}}{t}; \tensorSup{O}{t}\right] = \sigma\left(\mytensor{W} \ast \tensorSup{X}{t} + \tensorSup{V}{0}\right)\) \;
\(\tensorSup{C}{t} = \tensorSup{C}{t-1} + \tensorSup{\Tilde{C}}{t} \circ \tensorSup{I}{t}; ~ \tensorSup{H}{t} = \tensorSup{O}{t} \circ \sigma(\tensorSup{C}{t}) \) \;
return \(\tensorSup{H}{t}\), \(\tensorSup{C}{t}\)\;
\caption{{\bf Convolutional Tensor-Train LSTM} (Accelerated: \(T(N) = O(N)\)).}
\label{alg:conv-tt-lstm-fast}
\end{algorithm}}%
\section{Experimental Details}
\label{app:exp}

In this section, we provide the detailed setup of all experiments (datasets, model architectures, learning strategies and evaluation metrics) for both video prediction and early activity recognition.

\subsection{Preprocessing Module}
\label{app-sub:preprocessing}

In the main paper, we use a sliding window to group consecutive states in the proprocessing module (Section 3). In the Discussion (Section 5), we argued that other possible approaches are less effective in preserving spatio-temporal structure compared to our {\em sliding window approach}. Here, we discuss an alternative approach that was previously proposed for non-convolutional higher-order RNN~\citep{yu2017long}, which we name as {\em fixed window approach}. We will compare these two approaches in terms of computational complexity, ability to preserve temporal structure and predictive performance.

\begin{figure*}[!htbp]
  \centering
    \begin{subfigure}[b]{0.48\textwidth}
  \centering 
      \includegraphics[trim={0cm 0cm 0cm 0cm},clip, height=5cm]{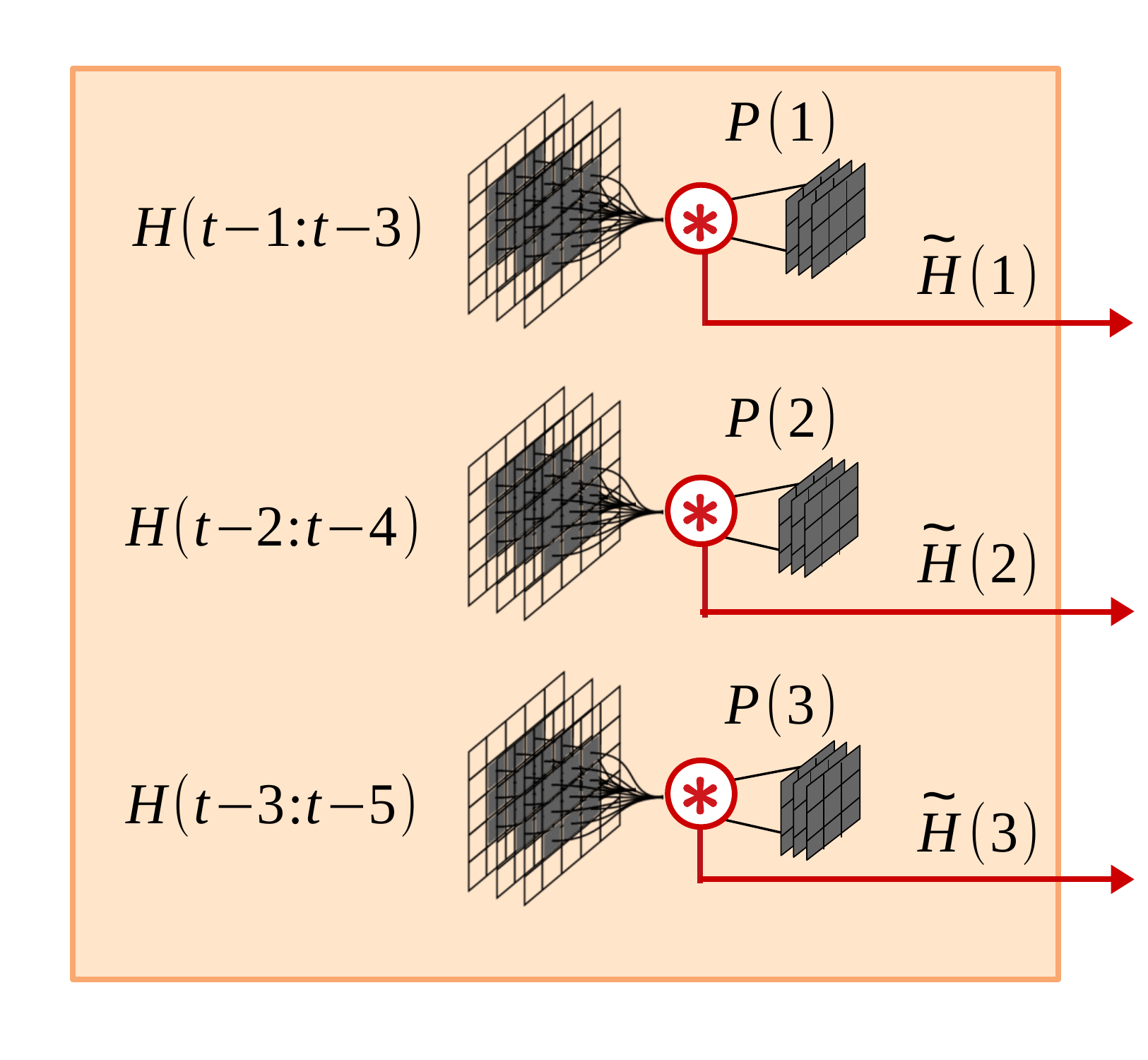}
      \caption{Sliding window approach (final implementation)}
      \label{subfig:sliding-window}
  \end{subfigure}
  \begin{subfigure}[b]{0.48\textwidth}
  \centering
       \includegraphics[trim={0cm 0cm 0cm 0.0cm},clip,height=5cm]{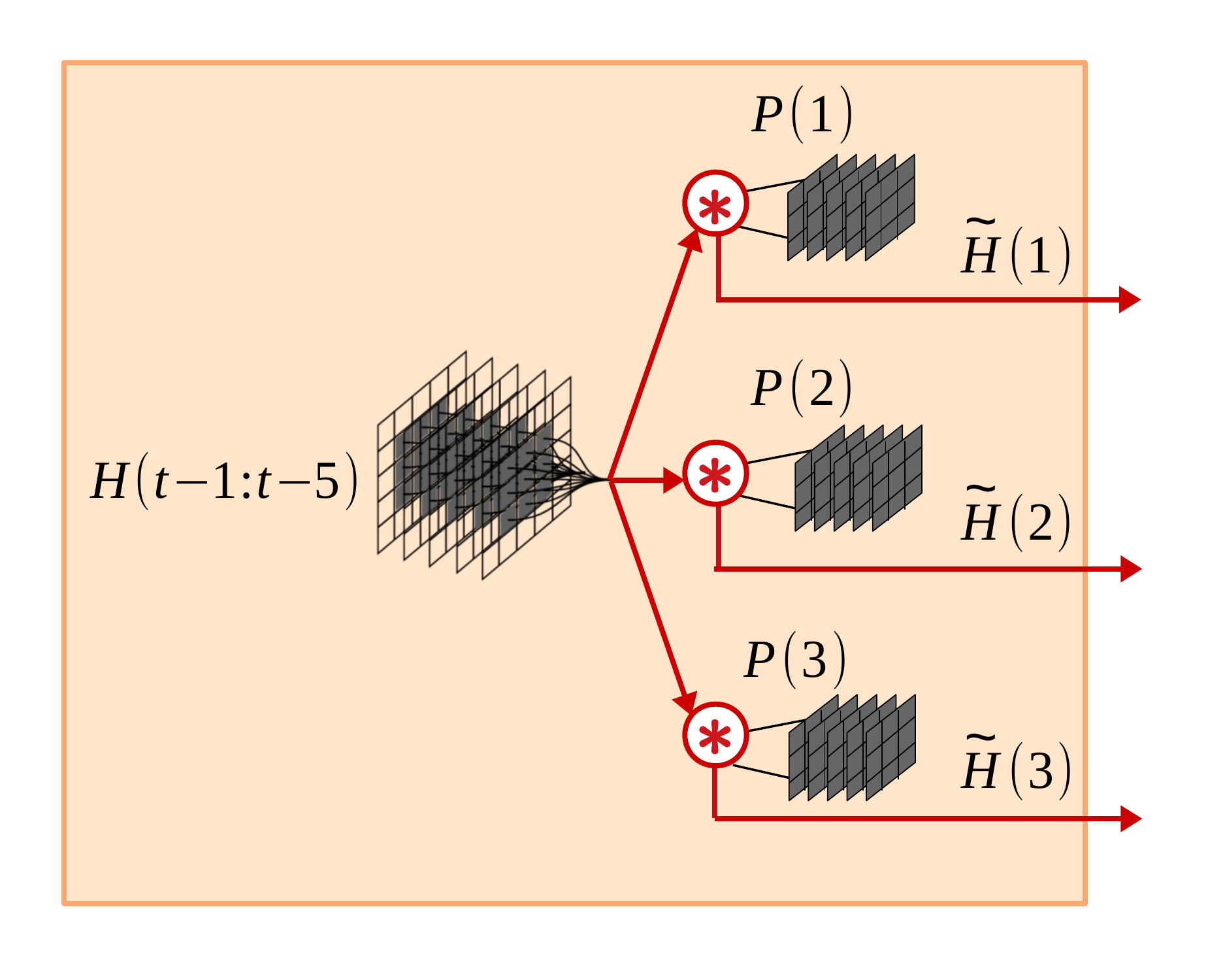}
      \caption{Fixed window approach (alternative)}
      \label{subfig:fixed-window}
  \end{subfigure}
  \caption{{\bf Variations of proprocessing modules.}}
   \label{fig:preprocessing}
\end{figure*}

\textbf{Fixed window approach.} 
With fixed window approach, $M$ previous steps $\{\tensorSup{H}{t-1}, \cdots, \tensorSup{H}{t-M}\}$ are first concatenated into a single tensor, which is then repeatedly mapped to $N$ inputs $\{\tensorSup{\Tilde{H}}{1}, \cdots, \tensorSup{\Tilde{H}}{N}\}$ to the $\mathsf{CTT}$ module.
\begin{subequations}
\begin{align}
\textbf{Fixed Window (FW): \quad}
\tensorSup{\tilde{H}}{i} 
& = 
\tensorSup{P}{i} \ast \left[\tensorSup{H}{t-1}; \cdots; \tensorSup{H}{t-N} \right] \\
\textbf{Sliding Window (SW): \quad}
\tensorSup{\tilde{H}}{i} 
& =
\tensorSup{P}{i} \ast \left[\tensorSup{H}{t-i}; \cdots; \tensorSup{H}{t-i+N-M} \right]
\end{align}
\end{subequations}
For comparison, we list both equations for fixed window approach and sliding window approach. These two approaches are also illustrated in \autoref{fig:preprocessing}.

\textbf{Drawbacks of fixed window approach.}
{\bf(a)} The {fixed window approach} has a larger window size than the {sliding window approach}, thus requires more parameters in the preprocessing kernels and higher computational complexity.
{\bf(b)} More importantly, the fixed window approach does not preserve the chronological order of the preprocessed states; unlike sliding window approach, the index \(i\) for \(\tensorSup{\Tilde{H}}{i}\) in fixed window approach cannot reflect the time for the compressed states. Actually, all preprocessed states \(\tensorSup{\Tilde{H}}{1}, \cdots, \tensorSup{\Tilde{H}}{M}\) are equivalent, which violates the property (2) in designing our convolutional tensor-train module (Section 3.1).
{\bf(c)} In \autoref{tab:eval-ablation-strategy}, we compare these two approaches on Moving-MNIST-2 under the same experimental setting, and we find that the sliding window approach performs slightly better than fixed window.
For all aforementioned reasons, we choose {sliding window approach} in our implementation of the preprocessing module.

\subsection{Model Architectures}
\label{app-sub:architecture}

\begin{figure*}[!htbp]
  \centering
  \begin{subfigure}{0.4\textwidth}
  \centering
      \includegraphics[trim={0cm 0cm 0cm 0.0cm},clip,height=5cm]{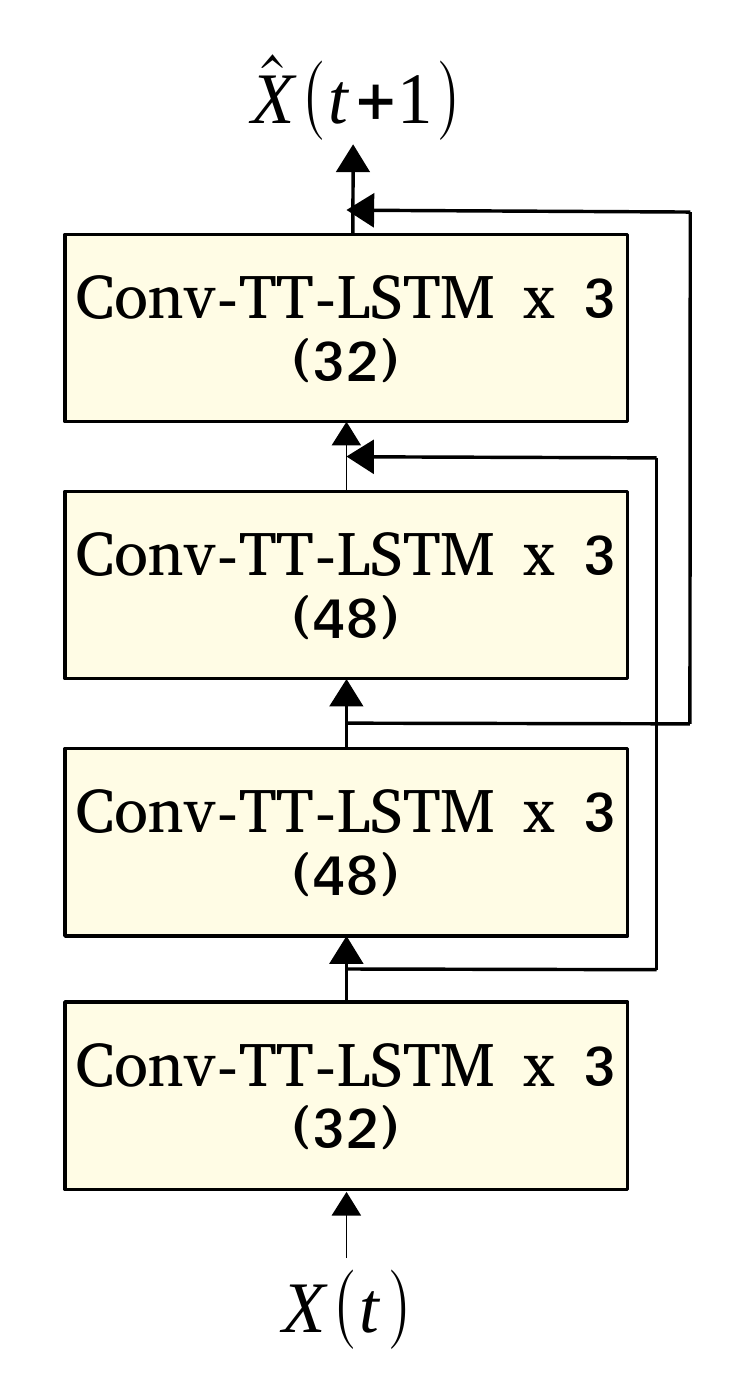}
      \caption{{\bf Prediction model}}
      \label{subfig:conv-tt-lstm-vp}
  \end{subfigure}
  \begin{subfigure}{0.40\textwidth}
  \centering 
     \vspace{0.3cm}
      \includegraphics[trim={0cm 0cm 0cm 0cm},clip, height=4.5cm]{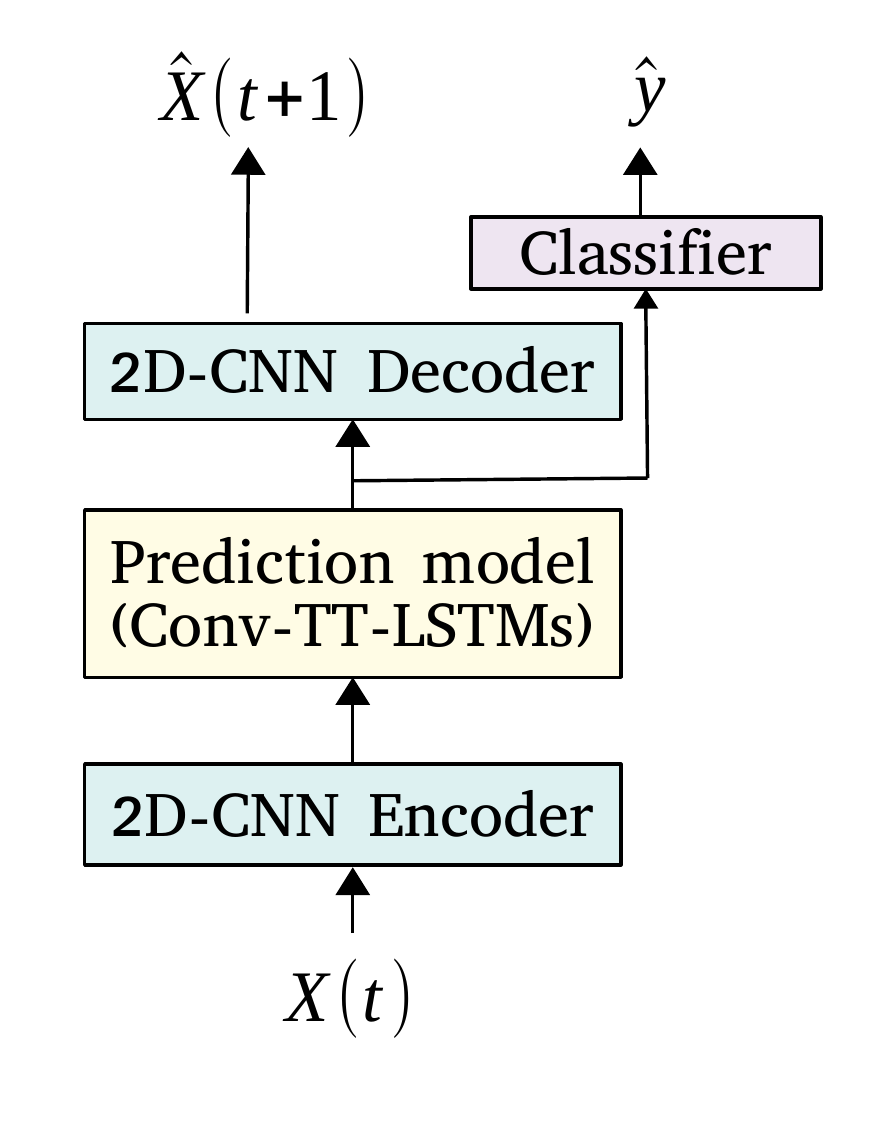}
      \vspace{0.2cm}
      \caption{{\bf Recognition model}}
      \label{subfig:conv-tt-lstm-vc}
  \end{subfigure}
  \caption{{\bf Network architecture for video prediction and early activity recognition tasks.}}
  \label{fig:arch}
\end{figure*}

\textbf{Video prediction.}
All experiments use a stack of 12-layers of {\ConvLSTM} or {\ConvTTLSTM} with 32 channels for the first and last 3 layers, and 48 channels for the 6 layers in the middle.
A convolutional layer is applied on top of all recurrent layers to compute the predicted frames, followed by an extra sigmoid layer for KTH action dataset.
Following~\citet{byeon2018contextvp}, two skip connections performing concatenation over channels are added between (3, 9) and (6, 12) layers.
An illustration of the network architecture is included in \autoref{subfig:conv-tt-lstm-vp}.
All convolutional kernels were initialized by Xavier's normalized initializer~\citep{glorot2010understanding} and initial hidden/cell states in {\ConvLSTM} or {\ConvTTLSTM} were initialized as zeros.

\textbf{Early activity recognition.}
Following~\citep{wang2018eidetic},
the network architecture consists of four modules: a 2D-CNN encoder, a video prediction network, a 2D-CNN decoder and a 3D-CNN classifier, as illustrated in \autoref{subfig:conv-tt-lstm-vc}. {\bf(1)} The 2D-CNN encoder has two $2$-strided 2D-convolutional layers with \(64\) channels, which reduce the resolution from \(224 \times 224\) to \(56 \times 56\), and {\bf(2)} the 2D-CNN decoder contains two $2$-strided transposed 2D-convolutional layers with \(64\) channels, which restore the resolution from \(56 \times 56\) to \(224 \times 224\). {\bf(3)} The video prediction network is miniature version of \autoref{subfig:conv-tt-lstm-vp}, where the number of layers in each block is reduced to $2$. In the experiments, we evaluate three realizations of each layer: {\ConvLSTM}, {\ConvTTLSTM} or causal 3D-convolutional layer.
{\bf(4)} The 3D-CNN classifier takes the last \(16\) frames from the input, and predicts a label for the \(41\) categories. The classifier contains two $2$-strided 3D-convolutional layers with \(128\) channels, each of which is followed by a 3D-pooling layer.
These layers reduce the resolution from \(56 \times 56\) to \(7 \times 7\), and the output feature is fed into a two-layers perceptron with \(512\) hidden units to compute the label.

\subsection{Hyper-parameters selection.}
\label{app-sub:hyperparameters}

\autoref{tab:hyperparameter} summarizes our search values for different hyper-parameters for {\ConvTTLSTM}.
{\bf(1)} For filter size \(K\), we found models with larger filter size \(K = 5\) consistently outperform the ones with \(K = 3\).
{\bf(2)} For learning rate, we found that our models are unstable at a high learning rate such as \(10^{-3}\), but learn poorly at a low learning rate $10^{-4}$. Consequently, we use gradient clipping with learning rate \(10^{-3}\), with clipping value $1$ for all experiments. {\bf(3)} While the performance typically increases as the order grows, the model suffers gradient instability in training with a high order, e.g.\ \(N = 5\). Therefore, we choose the order \(N = 3\) for all {\ConvTTLSTM} models. {\bf(4)(5)} For small ranks \(\scalarSup{C}{i}\) and steps \(M\), the performance increases monotonically with \(\scalarSup{C}{i}\) and \(M\). But the performance stays on plateau when we further increase them, therefore we settle down at \(\scalarSup{C}{i} = 8, \forall i\) and \(M = 5\) for all experiments.

\begin{table*}[!htbp]
  \centering
  \begin{tabular}{c c c c c}
    \toprule
     \footnotesize Filter size \(K\) &  \footnotesize Learning rate & \footnotesize Order of {\CTTD} \(N\) & \footnotesize Ranks of {\CTTD} \(\tensorSup{C}{i}\) & \footnotesize Time steps \(M\) \\
    \midrule
    \{3, 5\} & \{$10^{-4}$, $5 \times 10^{-4}$, $10^{-3}$\} & \{$1,2,3,5$\} & \{$4,8,16$\} & \{$1,3,5$\} \\
    \bottomrule
  \end{tabular}
   \caption{Hyper-parameters search values for Conv-TT-LSTM experiments.}
  \label{tab:hyperparameter}
\end{table*}

Similarly, \autoref{tab:hyperparameter-compression} summarize the hyper-parameters search for tensor-train compression of ConvLSTM~\citep{ garipov2016ultimate}. {\bf(1)} Since the best {\ConvLSTM} baseline has filter size \(K = 5\), we only consider \(K = 5\) in the compression experiments. {\bf(2)} We observe that the compressed {\ConvLSTM} models consistently achieve better performance with learning rate 
\(10^{-3}\). {\bf(3)(4)} The compressed {\ConvLSTM}s are robust to different order and ranks, and \(N = 2, R = 8\) wins by a small margin.

\begin{table*}[!htbp]
  \centering
  \begin{tabular}{c c c c}
    \toprule
     \footnotesize Filter size \(K\) &  \footnotesize Learning rate & \footnotesize Order of {\TTD} \(N\) & \footnotesize Ranks of {\TTD} \(R\)\\
    \midrule
    5 & \{$10^{-4}$, $10^{-3}$\} & \{$2,3$\} & \{$8,16,32$\} \\
    \bottomrule
  \end{tabular}
   \caption{Hyper-parameters search values for Tensor-Train compression of {\ConvLSTM}.}
  \label{tab:hyperparameter-compression}
\end{table*}

\subsection{Datasets}
\label{app-sub:dataset}

\textbf{Moving-MNIST-2 dataset.}
The Moving-MNIST-2 dataset is generated by moving two digits of size $28 \times 28$ in MNIST dataset within a $64 \times 64$ black canvas. These digits are placed at a random initial location, and move with constant velocity in the canvas and bounce when they reach the boundary.
Following~\citet{wang2018predrnnpp}, we generate 10,000 videos for training, 3,000 for validation, and 5,000 for test with default parameters in the generator\tablefootnote{The Python code for Moving-MNIST-2 generator is publicly available online in \cite{mnist-source}.}.

Similarly, we summarize the search values 
\textbf{KTH action dataset.}
The KTH action dataset~\cite{laptev2004recognizing} contains videos of 25 individuals performing 6 types of actions on a simple background.
Our experimental setup follows~\citet{wang2018predrnnpp}, which uses persons 1-16 for training and 17-25 for testing, and each frame is resized to $128 \times 128$ pixels.
All our models are trained to predict 10 frames given 10 input frames. During training, we randomly select 20 contiguous frames from the training videos as a sample and group every 10,000 samples into one epoch to apply the learning strategy as explained at the beginning of this section.

\textbf{Something-Something V2 dataset.}
The Something-Something V2 dataset~\citep{goyal2017something} is a benchmark for activity recognition, which can be download online\footnote{\url{https://20bn.com/datasets/something-something}}. Following~\citet{wang2018eidetic}, we use the official subset with \(41\) categories that contains \(55111\) training videos and \(7518\) test videos. The video length ranges between \(2\) and \(6\) seconds with \(24\) frames per second (fps). We reserve \(10\%\) of the training videos for validation, and use the remaining \(90\%\) for optimizing the models.

\subsection{Evaluation Metrics}
\label{app-sub:metric}

We use two traditional metrics MSE (or PSNR) and SSIM~\citep{wang2004image}, and a recently proposed deep-learning based metric LPIPS~\citep{zhang2018unreasonable}, which measures the similarity between deep features.
Since MSE (or PSNR) is based on pixel-wise difference, it favors vague and blurry predictions, which is not a proper measurement of perceptual similarity.
While SSIM was originally proposed to address the problem, \citet{zhang2018unreasonable} shows that their proposed LPIPS metric aligns better to human perception.

\subsection{Ablation Studies}
\label{app-sub:ablation}

Here, we show that our proposed {\ConvTTLSTM} consistently improves the performance of {\ConvLSTM}, regardless of the architecture, loss function and learning schedule used.
Specifically, we perform three ablation studies on our experimental setting, by \textbf{(1)} Reducing the number of layers from 12 layers to 4 layers (same as~\cite{xingjian2015convolutional} and~\cite{wang2018predrnnpp}); \textbf{(2)} Changing the loss function from $\mathcal{L}_1 + \mathcal{L}_2$ to $\mathcal{L}_1$ only; and \textbf{(3)} Disabling the scheduled sampling and use teacher forcing during training process.
We compare the performance of our proposed {\ConvTTLSTM} against the {\ConvLSTM} baseline in these ablated settings, \autoref{tab:eval-ablation-strategy}.
The results show that our proposed {\ConvTTLSTM} consistently outperforms {\ConvLSTM} in all settings, i.e.\ the {\ConvTTLSTM} model improves upon {\ConvLSTM} in a board range of setups, which is not limited to the certain setting used in our paper. These ablation studies further show that our setup is optimal for predictive learning in Moving-MNIST-2 dataset.

\begin{table*}[htbp]
\centering
\begingroup
\setlength{\tabcolsep}{5pt}
\renewcommand{\arraystretch}{1} 
    \begin{tabular}{c |c| c c c c c c | c c c | c}
    \toprule
    \multicolumn{2}{c|}{\multirow{2}{*}{Model}} & \multicolumn{2}{c}{Layers} & \multicolumn{2}{c}{Sched.} & \multicolumn{2}{c|}{Loss} & \multicolumn{3}{c|}{(10 -> 30)} & \multirow{2}{*}{Params.} \\    
    \multicolumn{2}{c|}{} & 4 & 12 & TF & SS & $\ell_1$ & $\ell_1 + \ell_2$ & MSE & SSIM & LPIPS & \\
    \midrule
    \midrule
    ConvLSTM & - & \multirow{2}{*}{\ding{51}} & \multirow{2}{*}{\ding{53}} & \multirow{2}{*}{\ding{53}} & \multirow{2}{*}{\ding{51}} & \multirow{2}{*}{\ding{53}} & \multirow{2}{*}{\ding{51}} & 37.19 & 0.791 & 184.2 & 11.48M \\
    Conv-TT-LSTM & FW & & & & & & & {\bf 31.46} & {\bf 0.819} & {\bf 112.5} &  \textbf{5.65M} \\
    \midrule
    ConvLSTM & - & \multirow{2}{*}{\ding{53}} & \multirow{2}{*}{\ding{51}} & \multirow{2}{*}{\ding{51}} & \multirow{2}{*}{\ding{53}} & \multirow{2}{*}{\ding{53}} & \multirow{2}{*}{\ding{51}} & 
    33.96 & 0.805 & 184.4 & 3.97M \\
    Conv-TT-LSTM & FW & & & & & & & 
    {\bf 30.27} & {\bf 0.827} & {\bf 118.2} & \textbf{2.65M}  \\
    \midrule
    ConvLSTM & - & \multirow{2}{*}{\ding{53}} & \multirow{2}{*}{\ding{51}} & \multirow{2}{*}{\ding{53}} & \multirow{2}{*}{\ding{51}} & \multirow{2}{*}{\ding{51}} & \multirow{2}{*}{\ding{53}} & 
    36.95 & 0.802 & 135.1 & 3.97M \\
    Conv-TT-LSTM & FW & & & & & & & 
    {\bf 34.84} & {\bf 0.807} & {\bf 128.4} &  \textbf{2.65M} \\
    \midrule
    \midrule
    ConvLSTM & - & \multirow{2}{*}{\ding{53}} & \multirow{2}{*}{\ding{51}} & \multirow{2}{*}{\ding{53}} & \multirow{2}{*}{\ding{51}} & \multirow{2}{*}{\ding{53}} & \multirow{2}{*}{\ding{51}} & 
    33.08 & 0.806 & 140.1 & 3.97M \\
    Conv-TT-LSTM & FW & & & & & & &
    {\bf 28.88} & {\bf 0.831} & {\bf 104.1} & \textbf{2.65M} \\
    \midrule
    Conv-TT-LSTM & SW & \ding{53} & \ding{51} & \ding{53} & \ding{51} & \ding{53} & \ding{51} & 25.81 & \textbf{0.840} & \textbf{90.38} & 2.69M \\
    \bottomrule
  \end{tabular}
  \endgroup
  \caption{Evaluation of {\ConvLSTM} and our {\ConvTTLSTM} under ablated settings. In this table, FW stands for {\em fixed window approach}, {SW} stands for {\em sliding window approach}; For learning scheduling, TF denotes {\em teaching forcing} and SS denotes {\em scheduled sampling}. The experiments show that \textbf{(1)} our {\ConvTTLSTM} is able to improve upon {\ConvLSTM} under all settings; \textbf{(2)} Our current learning approach is optimal in the search space; \textbf{(3)} The sliding window approach outperforms the fixed window one under the optimal experimental setting.}
  \label{tab:eval-ablation-strategy}
\end{table*}

\section{Additional Experimental Results}
\label{app:results}

\begin{figure*}[!htbp]
    \centering
    \begin{subfigure}{0.32\textwidth}
    \centering
    \includegraphics[trim={0.0cm 0 0.6cm 0.6cm},clip, width=\textwidth]{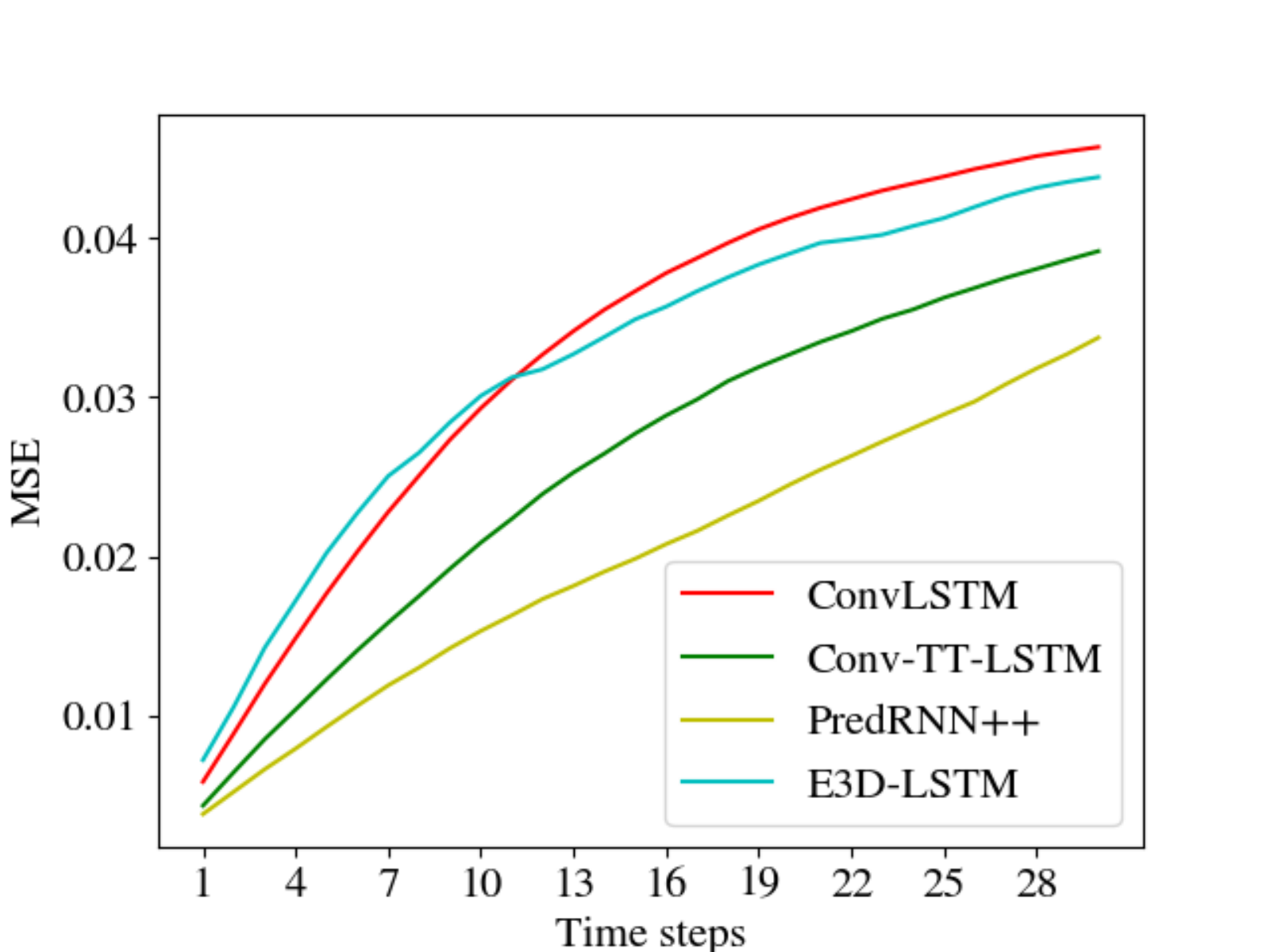}
    \end{subfigure}
    \hfill
    \begin{subfigure}{0.32\textwidth}
    \centering
    \includegraphics[trim={0.0cm 0 0.6cm 0.6cm},clip, width=\textwidth]{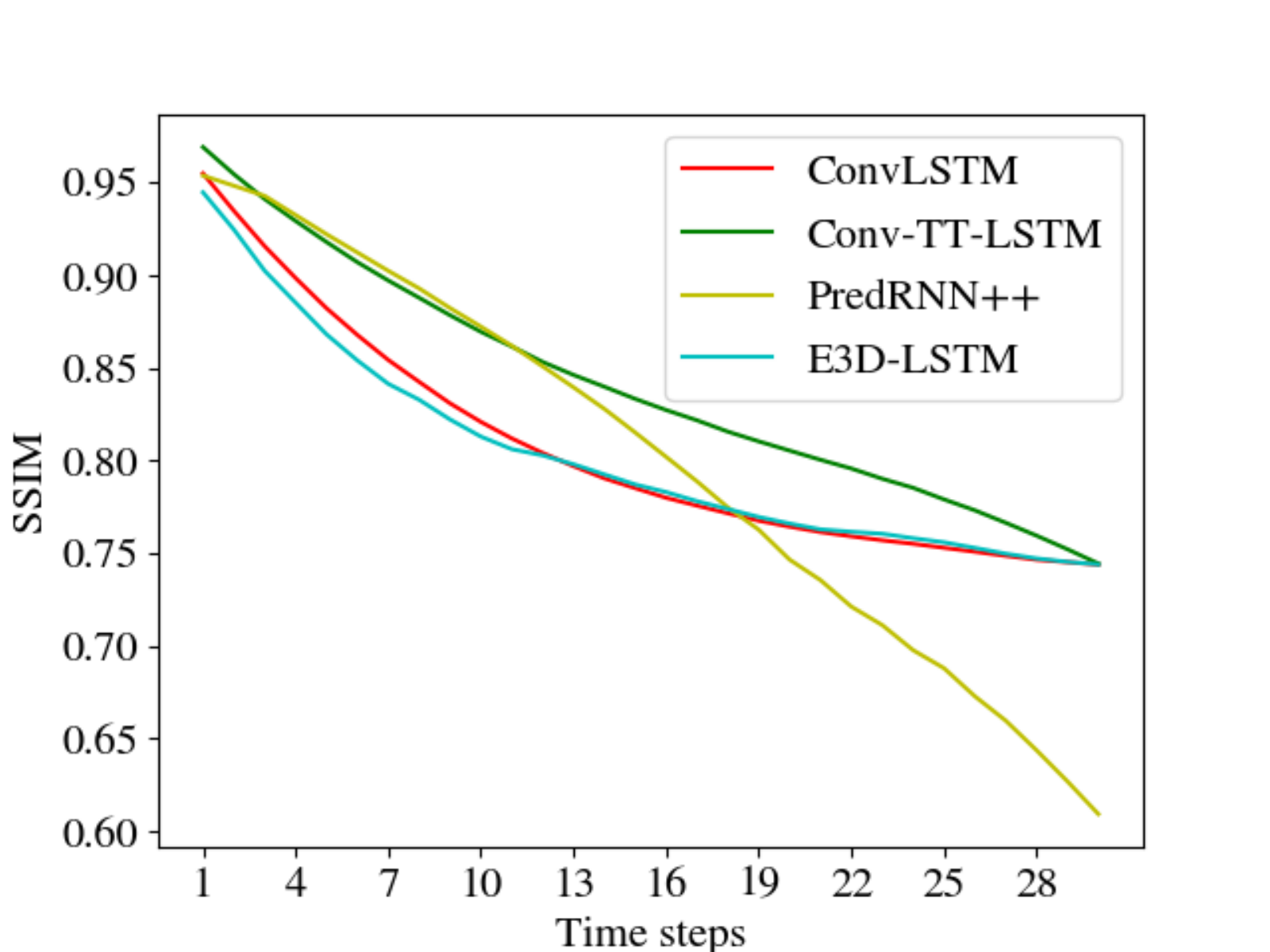}
  \end{subfigure}
  \begin{subfigure}{0.32\textwidth}
  \centering
      \includegraphics[trim={0.0cm 0 0.6cm 0.6cm},clip, width=\textwidth]{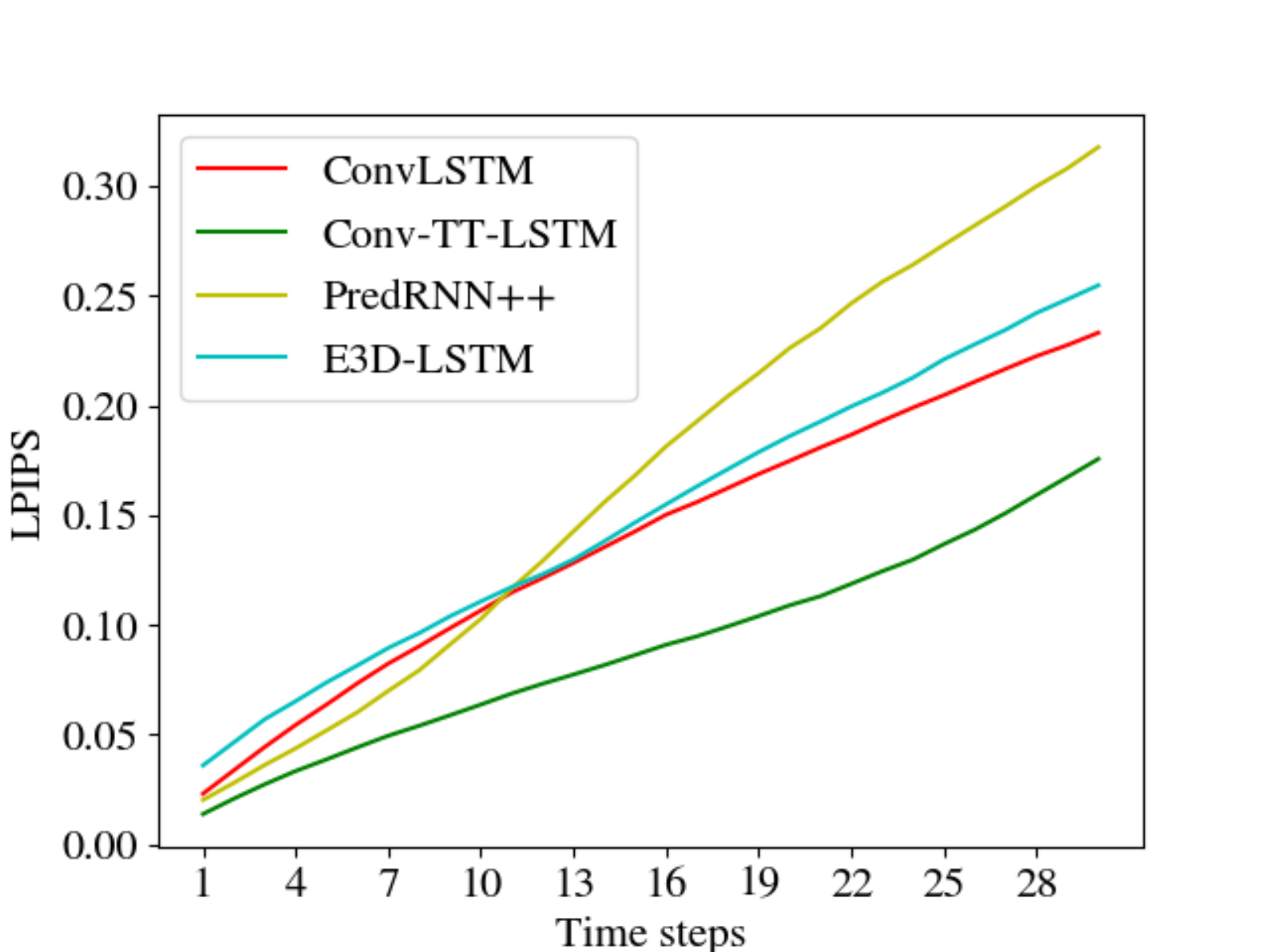}
  \end{subfigure}
  \caption{\textbf{Frame-wise comparison in MSE, SSIM and PIPS on Moving-MNIST-2 dataset.} For MSE and LPIPS, lower curves denote higher quality; while for SSIM, higher curves imply better quality. Our {\ConvTTLSTM} performs better than {\ConvLSTM} baseline, PredRNN++~\cite{wang2018predrnnpp} and E3D-LSTM ~\cite{wang2018eidetic} in all metrics (except for PredRNN++ in term of MSE).}
  \label{fig:mnist_per_frame_p30}
\end{figure*}

\begin{figure*}[!htbp]
  \centering
  \begin{subfigure}{0.32\textwidth}
  \centering
       \includegraphics[trim={0.0cm 0 0.6cm 0.6cm},clip, width=\textwidth]{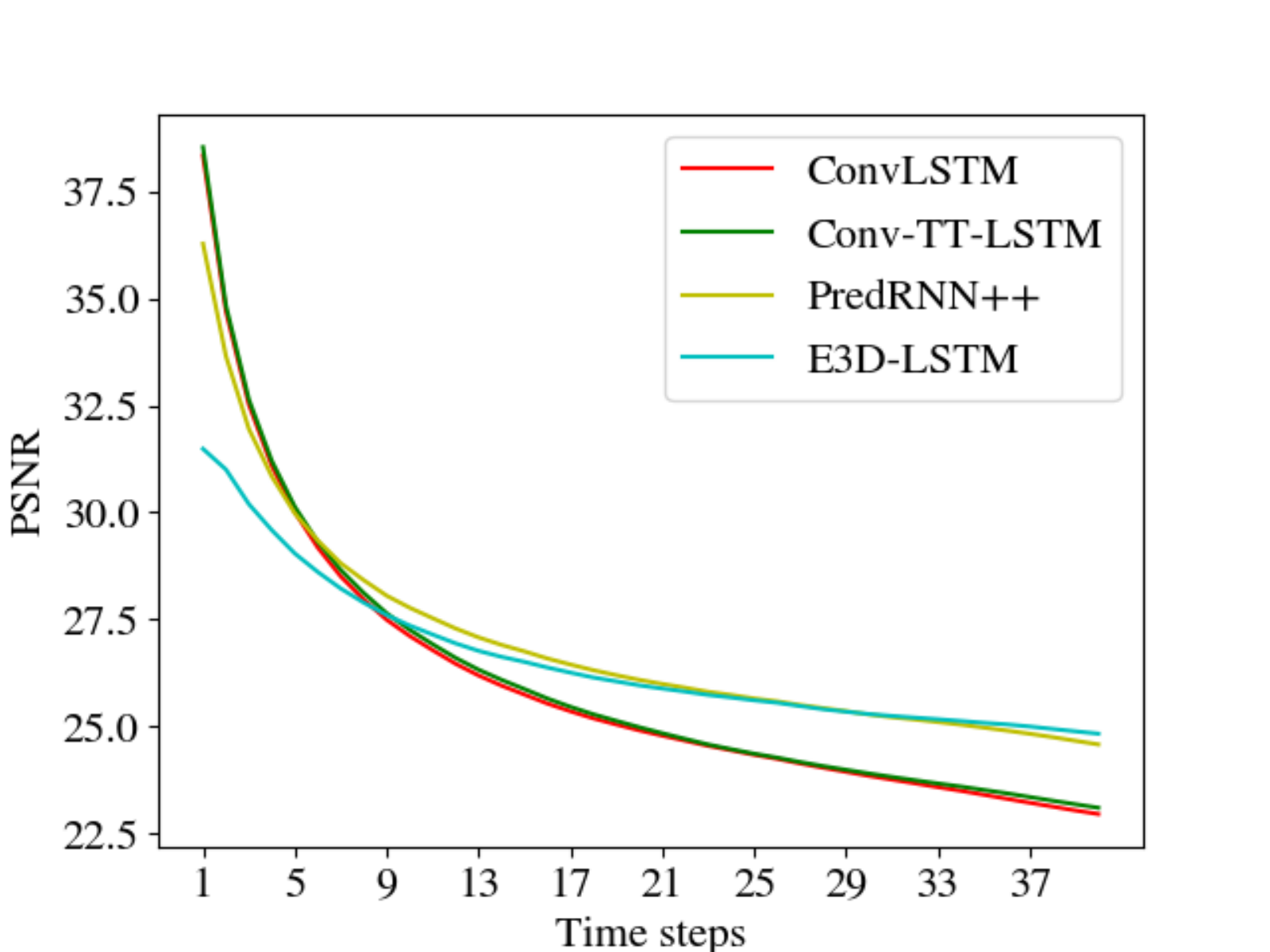}
  \end{subfigure}
  \begin{subfigure}{0.32\textwidth}
  \centering
       \includegraphics[trim={0.0cm 0 0.6cm 0.6cm},clip, width=\textwidth]{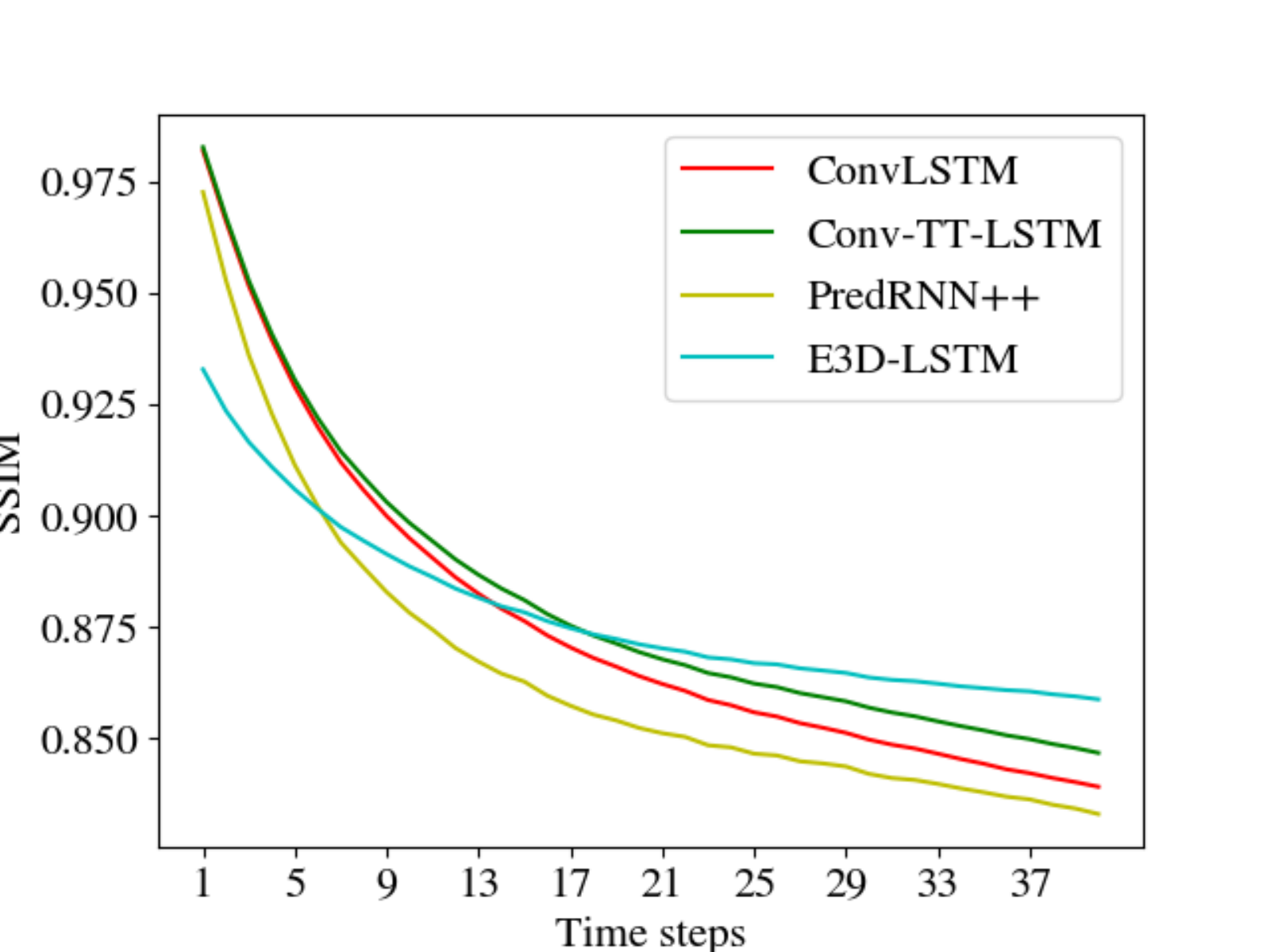}
  \end{subfigure}
  \begin{subfigure}{0.32\textwidth}
  \centering
       \includegraphics[trim={0.0cm 0 0.6cm 0.6cm},clip, width=\textwidth]{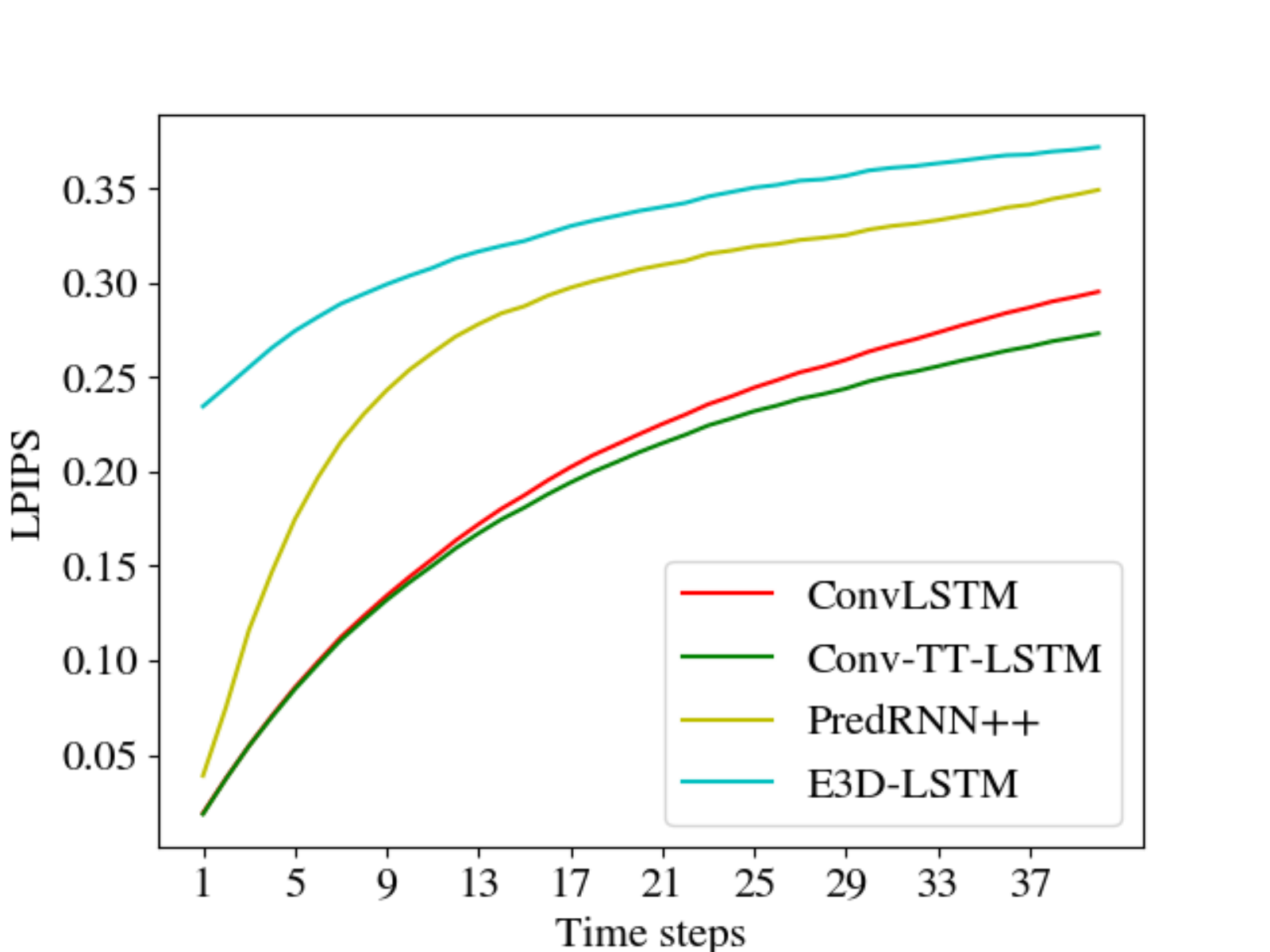}
  \end{subfigure}
  \caption{{\bf Frame-wise comparison in PSNR, SSIM and PIPS on KTH action dataset}. For LPIPS, lower curves denote higher quality; For PSNR and SSIM, higher curves imply better quality. Our {\ConvTTLSTM} outperforms {\ConvLSTM}, PredRNN++~\citep{wang2018predrnnpp} and E3D-LSTM~\citep{wang2018eidetic} in SSIM and LPIPS.}
  \label{fig:kth_per_frame_p40}
\end{figure*}

\textbf{Per-frame evaluations.} 
The per-frame metrics are illustrated in \autoref{fig:mnist_per_frame_p30} for Moving-MNIST-2 dataset, and \autoref{fig:kth_per_frame_p40} for KTH action dataset.
{\bf(1)} In Moving-MNIST-2 dataset, PredRNN++ performs comparably with our {\ConvTTLSTM} on early frames, but drops significantly for long-term prediction. E3D-LSTM performs similarly to {\ConvLSTM} baseline, and our {\ConvTTLSTM} consistently outperforms E3D-LSTM and {\ConvLSTM} over all frames.
{\bf(2)} In KTH action dataset, PredRNN++ consistently perform worse than our {\ConvTTLSTM} model over all frames; E3D-LSTM performs well on early frames in MSE and SSIM, but quickly deteriorates for long-term prediction.

\begin{figure*}[!htbp]
  \centering
  \addtolength{\tabcolsep}{-5pt}
  \resizebox{\textwidth}{!}{
  \begin{tabular}{c@{\hskip -0.003cm}cc@{\hskip -0.003cm}c}
      &input & & ground truth (top) / predictions \\ 
      \scriptsize $t = 1$ &\scriptsize $ 4\;\;\;\;\;\;\;\;\;6\;\;\;\;\;\;\;\;\;\;\;8$ &\scriptsize $11$ &  \scriptsize $14\;\;\;\;\;\;\;\;\;\;17\;\;\;\;\;\;\;\;\;20\;\;\;\;\;\;\;\;\;23\;\;\;\;\;\;\;\;\;\;\;26\;\;\;\;\;\;\;\;\;\;\;29\;\;\;\;\;\;\;\;\;\;\;32\;\;\;\;\;\;\;\;\;\;\;35\;\;\;\;\;\;\;\;\;\;\;38$ \\
      
      \animategraphics[height=0.95cm,loop,autoplay]{7}{figs/result_mnist_gif/ttv4_o3t3r8/inp_297_1632_}{0}{9} & 
      \includegraphics[trim={16.4cm 4.7cm 0.cm 0.1cm},clip, height=0.95cm]{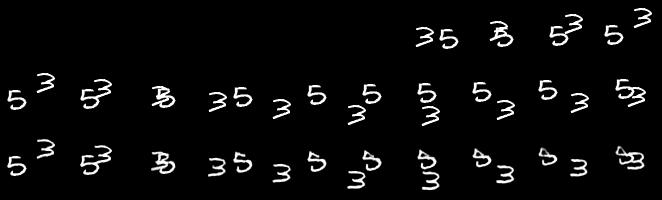} &
      \animategraphics[height=1.01cm,loop,autoplay]{7}{figs/result_mnist_gif/ttv4_o3t3r8/gt_297_1632_}{0}{29} & 
      \includegraphics[trim={2.5cm 2.5cm 0.0cm 2.4cm},clip, height=1cm,width=10cm]{figs/result_mnist/ttv4_o3t3r8/cmp_297_1632.jpg} \\ [-0.2em]
      
      & \raisebox{0.4cm}{PredRNN++} &
      \animategraphics[height=1.01cm,loop,autoplay]{7}{figs/result_mnist_gif/predrnnpp/pred_160000_1648_}{0}{29} &
      \includegraphics[trim={2.5cm 0 0.cm 4.7cm},clip, height=1cm,width=10cm]{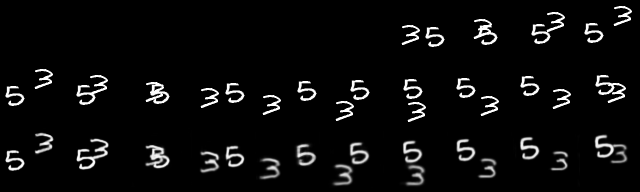} \\
      [-0.2em]
      
      & \raisebox{0.4cm}{E3D-LSTM} &
      \animategraphics[height=1.01cm,loop,autoplay]{7}{figs/result_mnist_gif/e3d/pred_80000_1648_}{0}{29} &
      \includegraphics[trim={2.5cm 0 0.cm 4.7cm},clip, height=1cm,width=10cm]{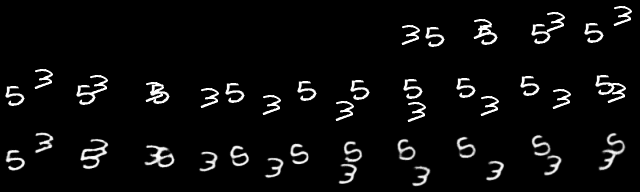} \\ [-0.2em]
      
      & \raisebox{0.4cm}{{\ConvLSTM}} &
      \animategraphics[height=1.01cm,loop,autoplay]{7}{figs/result_mnist_gif/baseline/pred_278_1632_}{0}{29} &
      \includegraphics[trim={2.5cm 0 0.cm 4.7cm},clip, height=1cm,width=10cm]{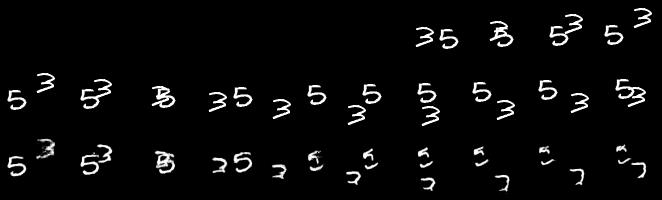} \\ [-0.2em]
      
      & \raisebox{0.4cm}{{\ConvTTLSTM}} &
      \animategraphics[height=1.01cm,loop,autoplay]{7}{figs/result_mnist_gif/ttv4_o3t3r8/pred_297_1632_}{0}{19} &
      \includegraphics[trim={2.5cm 0 0.cm 4.7cm},clip, height=1cm,width=10cm]{figs/result_mnist/ttv4_o3t3r8/cmp_297_1632.jpg} \\ [-0.2em]
  \end{tabular}}
  \caption{30 frames prediction on Moving-MNIST given 10 input frames. Every 3 frames are shown. The first frames ($t=1$ and $11$) are animations. To view the animation, Adobe reader is required.}
  \label{fig:visual-mnist1}
\end{figure*}

\begin{figure*}[!htbp]
  \centering
  \addtolength{\tabcolsep}{-5pt}
  \resizebox{\textwidth}{!}{
  \begin{tabular}{c@{\hskip -0.003cm}cc@{\hskip -0.003cm}c}
      &input & & ground truth (top) / predictions \\ 
      \scriptsize $t = 1$ &\scriptsize $ 4\;\;\;\;\;\;\;\;\;6\;\;\;\;\;\;\;\;\;\;\;8$ &\scriptsize $11$ &  \scriptsize $14\;\;\;\;\;\;\;\;\;\;17\;\;\;\;\;\;\;\;\;20\;\;\;\;\;\;\;\;\;23\;\;\;\;\;\;\;\;\;\;\;26\;\;\;\;\;\;\;\;\;\;\;29\;\;\;\;\;\;\;\;\;\;\;32\;\;\;\;\;\;\;\;\;\;\;35\;\;\;\;\;\;\;\;\;\;\;38$ \\
      
      \animategraphics[height=1.01cm,loop,autoplay]{7}{figs/result_mnist_gif/ttv4_o3t3r8/inp_297_2400_}{0}{9} & 
      \includegraphics[trim={16.4cm 4.7cm 0.cm 0.1cm},clip, height=1cm]{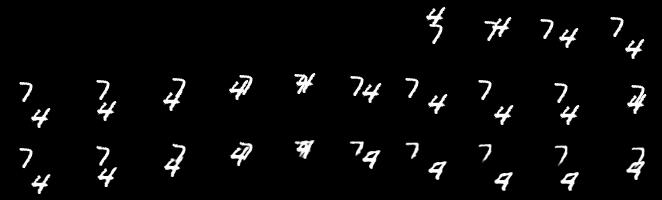} &
      \animategraphics[height=1.01cm,loop,autoplay]{7}{figs/result_mnist_gif/ttv4_o3t3r8/gt_297_2400_}{0}{29} & 
      \includegraphics[trim={2.5cm 2.5cm 0.0cm 2.4cm},clip, height=1cm,width=10cm]{figs/result_mnist/ttv4_o3t3r8/cmp_297_2400.jpg} \\ [-0.2em]
      
      & \raisebox{0.4cm}{PredRNN++} &
      \animategraphics[height=1.01cm,loop,autoplay]{7}{figs/result_mnist_gif/predrnnpp/pred_160000_2416_}{0}{29} &
      \includegraphics[trim={2.5cm 0 0.cm 4.7cm},clip, height=1cm,width=10cm]{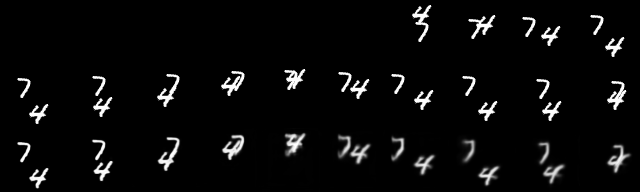} \\ [-0.2em]
      
      & \raisebox{0.4cm}{E3D-LSTM} &
      \animategraphics[height=1.01cm,loop,autoplay]{7}{figs/result_mnist_gif/e3d/pred_80000_2416_}{0}{29} &
      \includegraphics[trim={2.5cm 0 0.cm 4.7cm},clip, height=1cm,width=10cm]{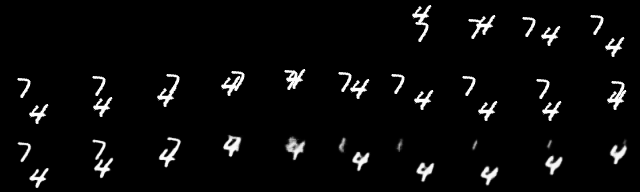} \\ [-0.2em]
      
      & \raisebox{0.4cm}{{\ConvLSTM}} &
      \animategraphics[height=1.01cm,loop,autoplay]{7}{figs/result_mnist_gif/baseline/pred_278_2400_}{0}{29} &
      \includegraphics[trim={2.5cm 0 0.cm 4.7cm},clip, height=1cm,width=10cm]{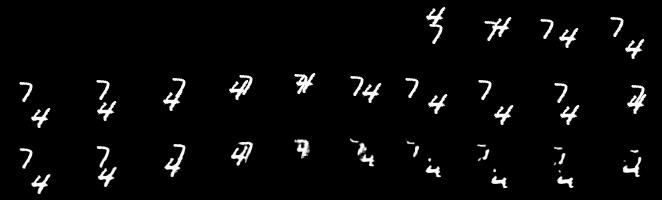} \\ [-0.2em]
      
      & \raisebox{0.4cm}{{\ConvTTLSTM}} &
      \animategraphics[height=1.01cm,loop,autoplay]{7}{figs/result_mnist_gif/ttv4_o3t3r8/pred_297_2400_}{0}{19} &
      \includegraphics[trim={2.5cm 0 0.cm 4.7cm},clip, height=1cm,width=10cm]{figs/result_mnist/ttv4_o3t3r8/cmp_297_2400.jpg} \\ [-0.2em]
  \end{tabular}}
  \caption{30 frames prediction on Moving-MNIST given 10 input frames. Every 3 frames are shown. The first frames ($t=1$ and $11$) are animations. To view the animation, Adobe reader is required.}
  \label{fig:visual-mnist2}
\end{figure*}

\textbf{Additional visual results: Video prediction.}
\autoref{fig:visual-kth1}, \ref{fig:visual-kth2}, \ref{fig:visual-kth3}, \ref{fig:visual-kth4}, \ref{fig:visual-mnist1}, and \ref{fig:visual-mnist2} show additional visual comparisons.
We also attach two video clips (KTH and MNIST) as supplementary material. 

\textbf{Additional visual results: Early activity recognition.}
We attach two video clips (video 1 and 2) as supplementary material. The videos show the comparisons among 3D-CNN, Conv-LSTM and our Conv-TT-LSTM when the input frames are partially seen. The time-frame of the video corresponds to an amount of video frames seen by the models. 

\def\i{483}
\def\j{1323}
\def\ia{483}
\def\ja{1323}
\begin{figure*}[!htbp]
  \centering
  \footnotesize
  \addtolength{\tabcolsep}{-5pt}
  \resizebox{\textwidth}{!}{
  \begin{tabular}{cc}
      input & ground truth (top) / predictions \\ 
      \scriptsize $t = 1 \;\;\;\;\;\;\;\;4\;\;\;\;\;\;\;\;\;\;\;6\;\;\;\;\;\;\;\;\;\;8\;\;\;\;\;\;$ &\scriptsize $11\;\;\;\;\;\;\;\;13\;\;\;\;\;\;\;\;\;\;15\;\;\;\;\;\;\;\;\;17\;\;\;\;\;\;\;\;19\;\;\;\;\;\;\;\;21\;\;\;\;\;\;\;\;\;\;23\;\;\;\;\;\;\;\;\;25\;\;\;\;\;\;\;\;\;\;27\;\;\;\;\;\;\;\;29$ \\
      
      \includegraphics[trim={23cm 9.2cm 4.5cm 0.1cm},clip, height=1cm]{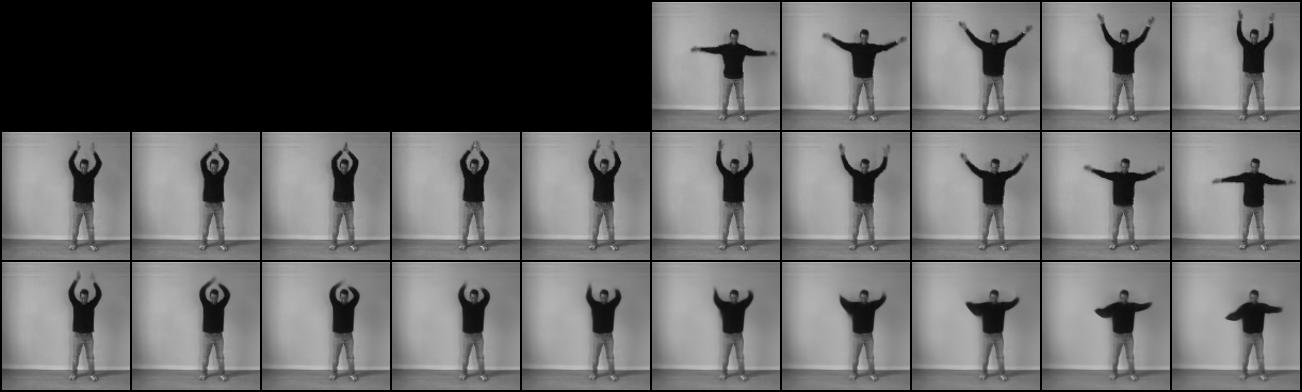} &
      \includegraphics[trim={0cm 4.7cm 0.0cm 4.8cm},clip, height=1cm,width=10cm]{figs/result_kth/ttv4_o3t3r8_ploss/cmp_395_\i.jpg} \\ [-0.2em]

      \raisebox{0.4cm}{E3D-LSTM} &
      \includegraphics[trim={0cm 0 0.cm 9.0cm},clip, height=1cm,width=10cm]{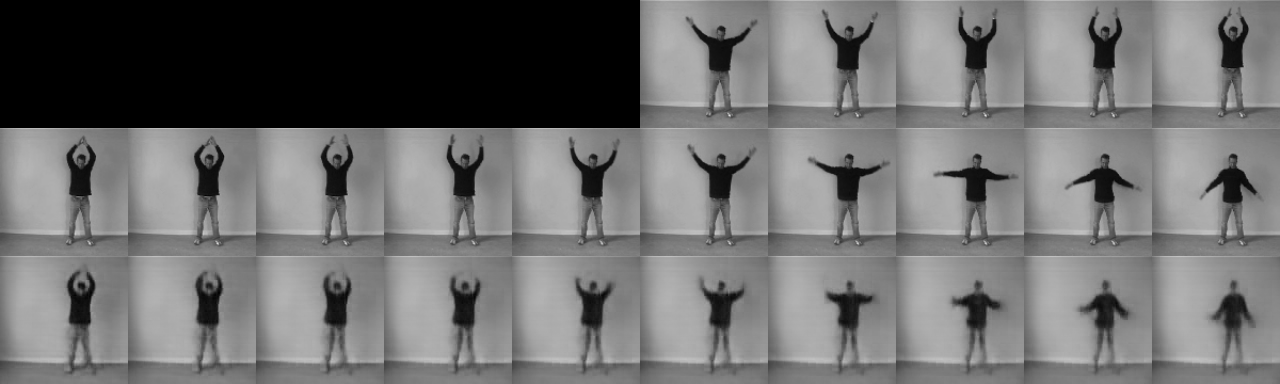} \\ [-0.2em]

      \raisebox{0.4cm}{ \ConvTTLSTM } &
      \includegraphics[trim={0cm 0.1cm 0.0cm 9.3cm},clip, height=1cm,width=10cm]{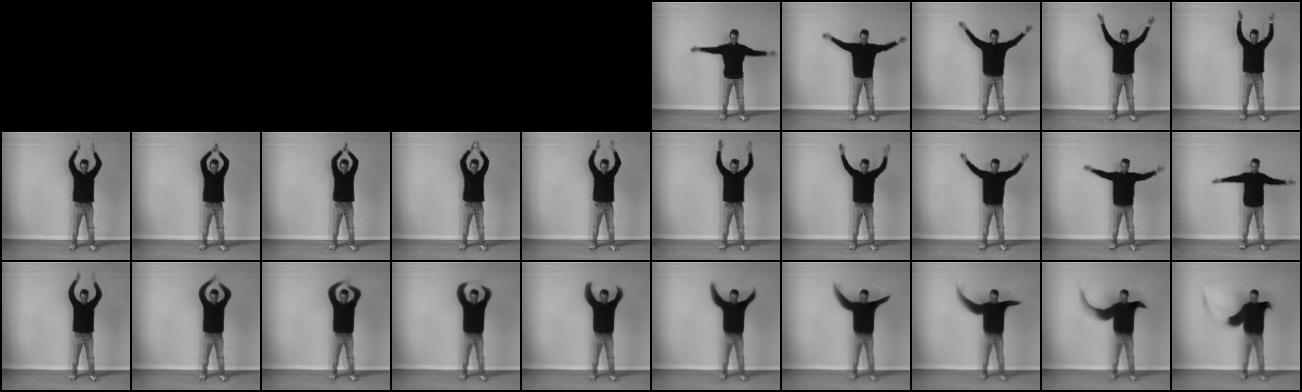} \\ [-0.2em]


  \end{tabular}}
  \caption{20 frames prediction on KTH given 10 input frames. Every 2 frames are shown.}
  \label{fig:visual-kth1}
  \vspace{-0.2cm}
\end{figure*}

\def\i{844}
\def\j{741}
\def\ia{844}
\def\ja{741}
\begin{figure*}[!htbp]
  \centering
  \footnotesize
  \addtolength{\tabcolsep}{-5pt}
  \resizebox{\textwidth}{!}{
  \begin{tabular}{cc}
      input & ground truth (top) / predictions \\ 
      \scriptsize $t = 1 \;\;\;\;\;\;\;\;4\;\;\;\;\;\;\;\;\;\;\;6\;\;\;\;\;\;\;\;\;\;8\;\;\;\;\;\;$ &\scriptsize $11\;\;\;\;\;\;\;\;13\;\;\;\;\;\;\;\;\;\;15\;\;\;\;\;\;\;\;\;17\;\;\;\;\;\;\;\;19\;\;\;\;\;\;\;\;21\;\;\;\;\;\;\;\;\;\;23\;\;\;\;\;\;\;\;\;25\;\;\;\;\;\;\;\;\;\;27\;\;\;\;\;\;\;\;29$ \\
      
      \includegraphics[trim={23cm 9.2cm 4.5cm 0.1cm},clip, height=1cm]{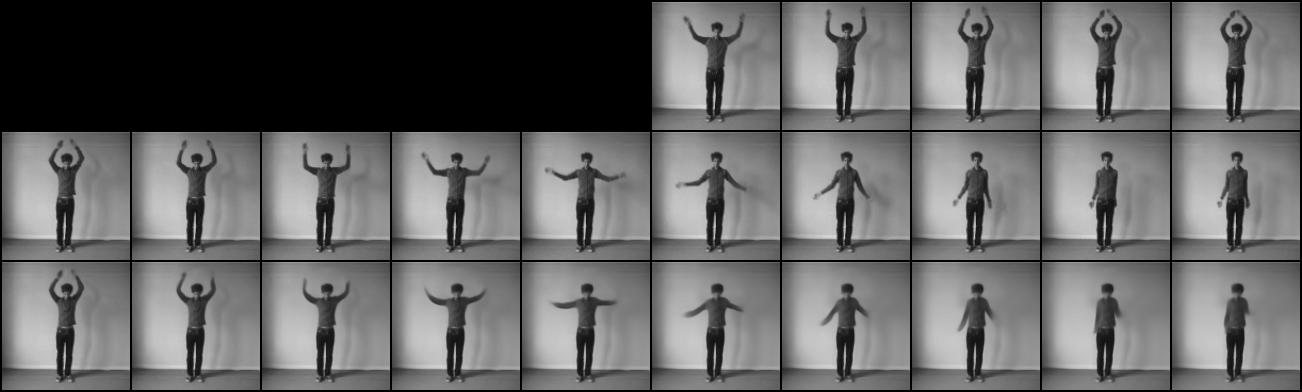} &
      \includegraphics[trim={0cm 4.7cm 0.0cm 4.8cm},clip, height=1cm,width=10cm]{figs/result_kth/ttv4_o3t3r8_ploss/cmp_395_\i.jpg} \\ [-0.2em]

      \raisebox{0.4cm}{E3D-LSTM} &
      \includegraphics[trim={0cm 0 0.cm 9.0cm},clip, height=1cm,width=10cm]{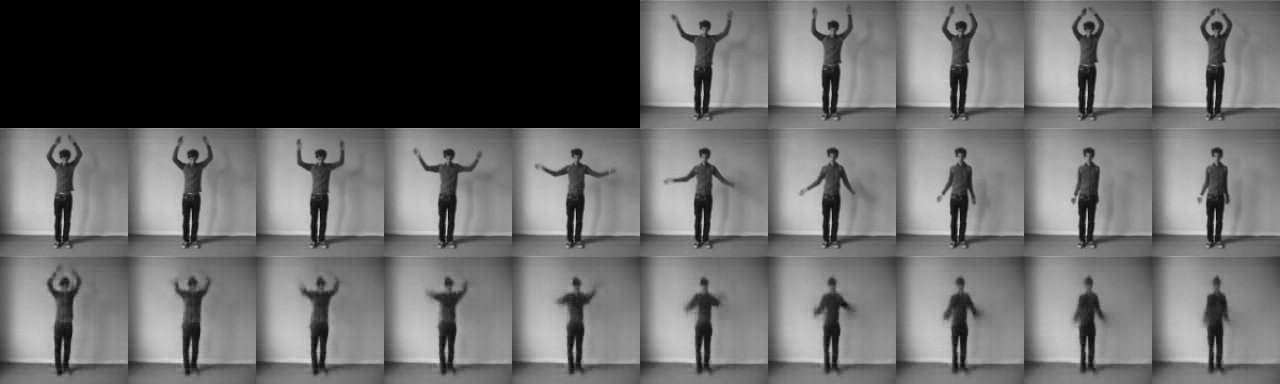} \\ [-0.2em]

      \raisebox{0.4cm}{ \ConvTTLSTM } &
      \includegraphics[trim={0cm 0.1cm 0.0cm 9.3cm},clip, height=1cm,width=10cm]{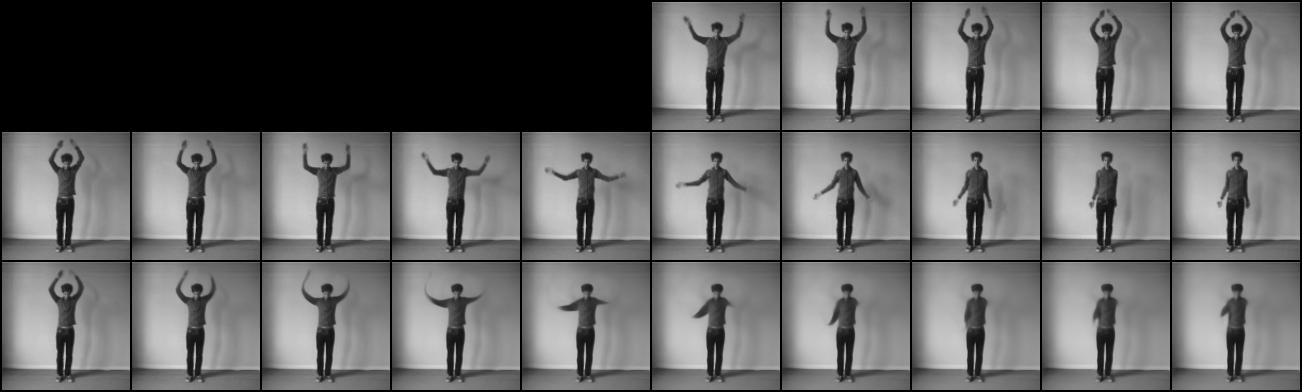} \\ [-0.2em]


  \end{tabular}}
  \caption{20 frames prediction on KTH given 10 input frames. Every 2 frames are shown.}
  \label{fig:visual-kth2}
  \vspace{-0.2cm}
\end{figure*}

\def\i{466}
\def\j{3108}
\def\ia{466}
\def\ja{3108}
\begin{figure*}[!htbp]
  \centering
  \footnotesize
  \addtolength{\tabcolsep}{-5pt}
  \resizebox{\textwidth}{!}{
  \begin{tabular}{cc}
      input & ground truth (top) / predictions \\ 
      \scriptsize $t = 1 \;\;\;\;\;\;\;\;4\;\;\;\;\;\;\;\;\;\;\;6\;\;\;\;\;\;\;\;\;\;8\;\;\;\;\;\;$ &\scriptsize $11\;\;\;\;\;\;\;\;13\;\;\;\;\;\;\;\;\;\;15\;\;\;\;\;\;\;\;\;17\;\;\;\;\;\;\;\;19\;\;\;\;\;\;\;\;21\;\;\;\;\;\;\;\;\;\;23\;\;\;\;\;\;\;\;\;25\;\;\;\;\;\;\;\;\;\;27\;\;\;\;\;\;\;\;29$ \\
      
      \includegraphics[trim={23cm 9.2cm 4.5cm 0.1cm},clip, height=1cm]{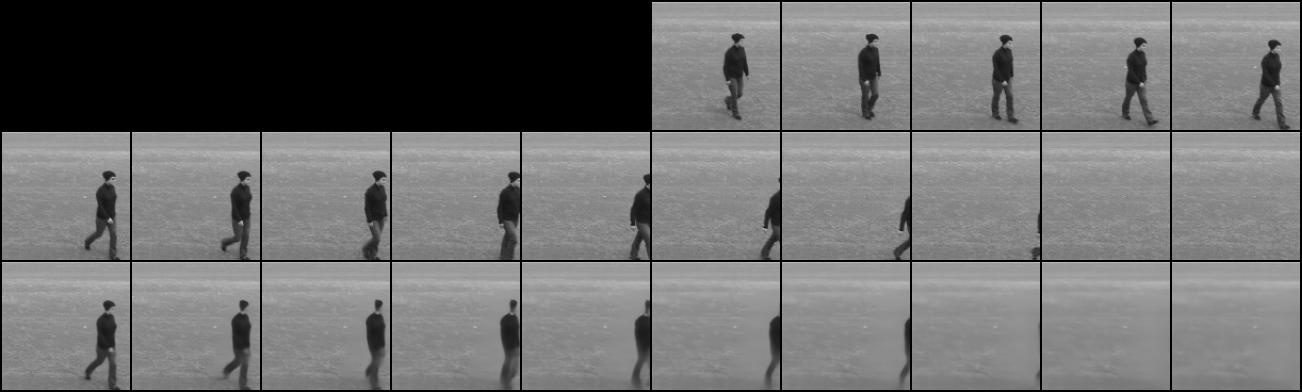} &
      \includegraphics[trim={0cm 4.7cm 0.0cm 4.8cm},clip, height=1cm,width=10cm]{figs/result_kth/ttv4_o3t3r8_ploss/cmp_395_\i.jpg} \\ [-0.2em]

      \raisebox{0.4cm}{E3D-LSTM} &
      \includegraphics[trim={0cm 0 0.cm 9.0cm},clip, height=1cm,width=10cm]{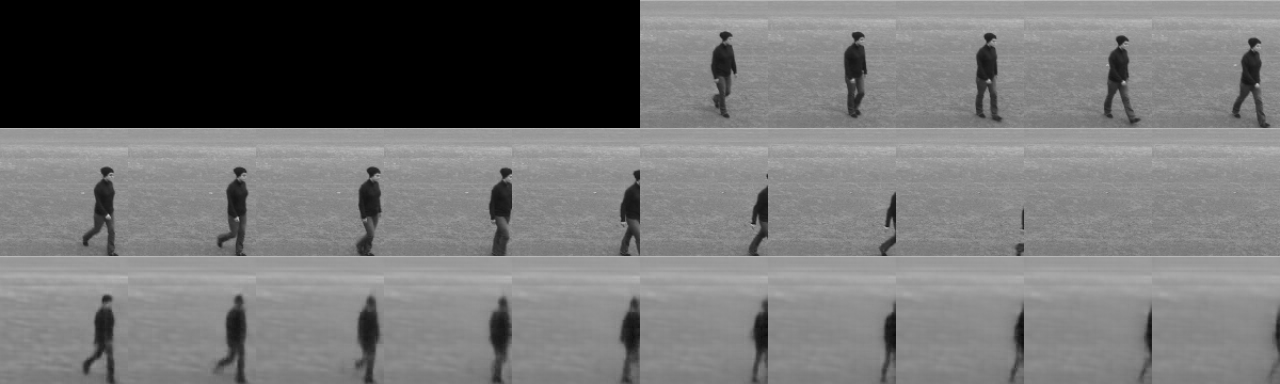} \\ [-0.2em]

      \raisebox{0.4cm}{ \ConvTTLSTM } &
      \includegraphics[trim={0cm 0.1cm 0.0cm 9.3cm},clip, height=1cm,width=10cm]{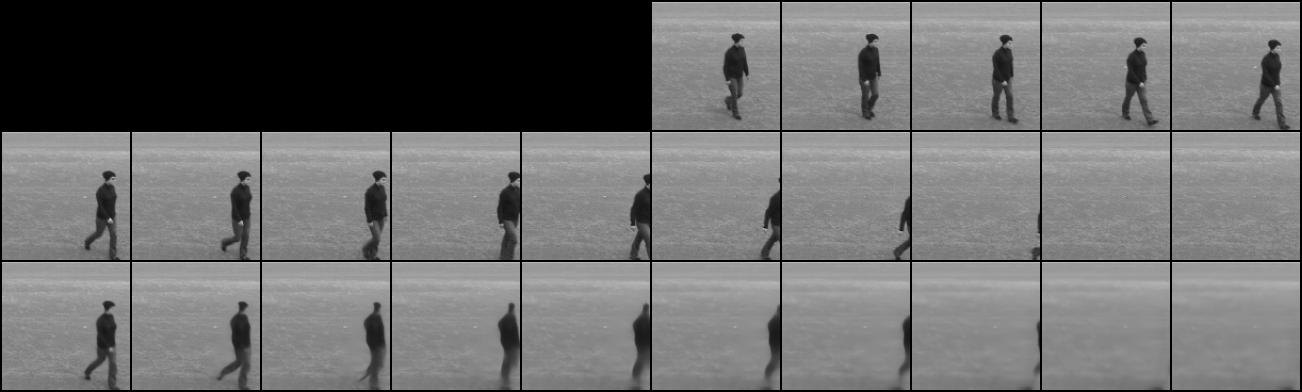} \\ [-0.2em]


  \end{tabular}}
  \caption{20 frames prediction on KTH given 10 input frames. Every 2 frames are shown.}
  \label{fig:visual-kth3}
  \vspace{-0.2cm}
\end{figure*}

\def\i{692}
\def\j{695}
\def\ia{692}
\def\ja{695}
\begin{figure*}[!htbp]
  \centering
  \footnotesize
  \addtolength{\tabcolsep}{-5pt}
  \resizebox{\textwidth}{!}{
  \begin{tabular}{cc}
      input & ground truth (top) / predictions \\ 
      \scriptsize $t = 1 \;\;\;\;\;\;\;\;4\;\;\;\;\;\;\;\;\;\;\;6\;\;\;\;\;\;\;\;\;\;8\;\;\;\;\;\;$ &\scriptsize $11\;\;\;\;\;\;\;\;13\;\;\;\;\;\;\;\;\;\;15\;\;\;\;\;\;\;\;\;17\;\;\;\;\;\;\;\;19\;\;\;\;\;\;\;\;21\;\;\;\;\;\;\;\;\;\;23\;\;\;\;\;\;\;\;\;25\;\;\;\;\;\;\;\;\;\;27\;\;\;\;\;\;\;\;29$ \\
      \includegraphics[trim={23cm 9.2cm 4.5cm 0.1cm},clip, height=1cm]{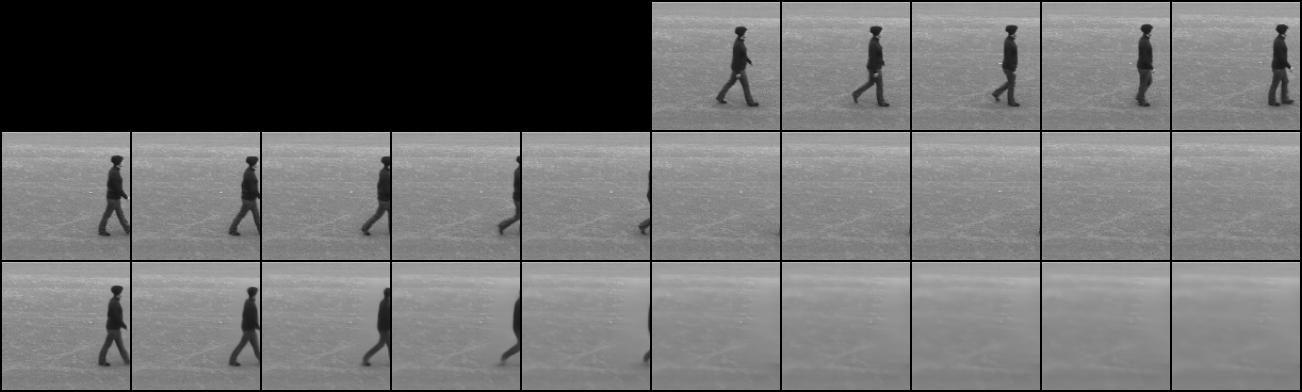} &
      \includegraphics[trim={0cm 4.7cm 0.0cm 4.8cm},clip, height=1cm,width=10cm]{figs/result_kth/ttv4_o3t3r8_ploss/cmp_395_\i.jpg} \\ [-0.2em]

      \raisebox{0.4cm}{E3D-LSTM} &
      \includegraphics[trim={0cm 0 0.cm 9.0cm},clip, height=1cm,width=10cm]{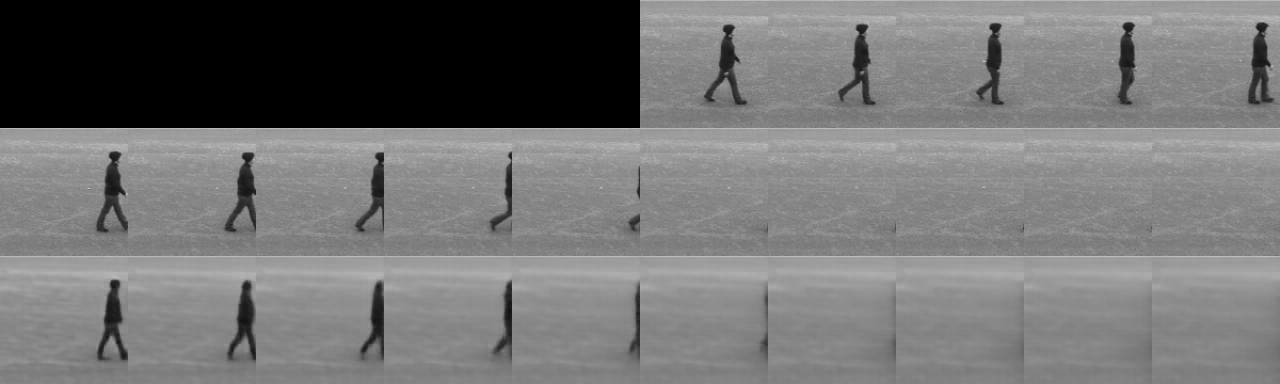} \\ [-0.2em]

      \raisebox{0.4cm}{ \ConvTTLSTM } &
      \includegraphics[trim={0cm 0.1cm 0.0cm 9.3cm},clip, height=1cm,width=10cm]{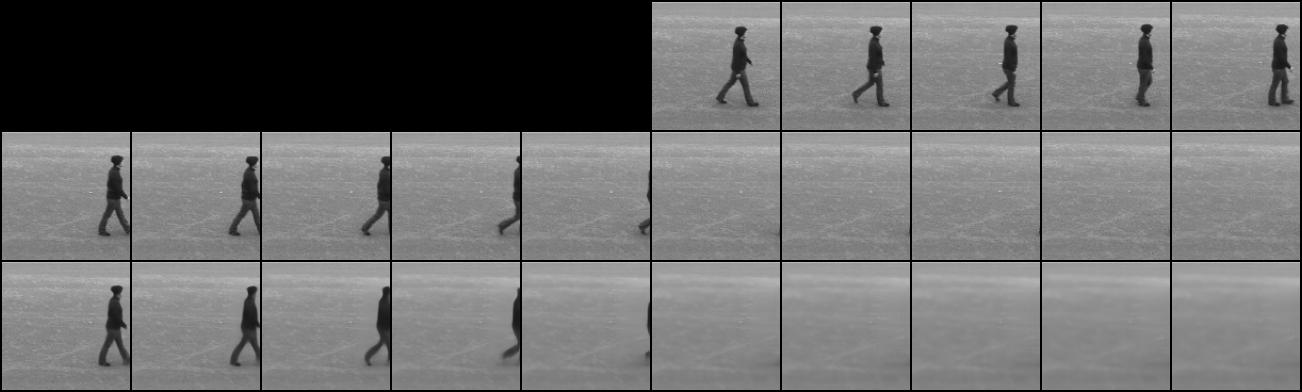} \\ [-0.2em]

  \end{tabular}}
  \caption{20 frames prediction on KTH given 10 input frames. Every 2 frames are shown. }
  \label{fig:visual-kth4}
  \vspace{-0.2cm}
\end{figure*}

\end{document}